\documentclass[journal]{IEEEtran}
\usepackage{amsmath,amsfonts}
\usepackage{algorithmic}
\usepackage{algorithm}
\usepackage{array}
\usepackage{colortbl}
\usepackage{color}
\usepackage[table,xcdraw]{xcolor}
\usepackage{textcomp}
\usepackage{stfloats}
\usepackage{url}
\usepackage{verbatim}
\usepackage{newclude}
\usepackage{graphicx}
\usepackage{cite}
\usepackage{makecell}
\usepackage{bm}

\usepackage[caption=false, font=footnotesize, labelfont=rm, textfont=rm]{subfig}
\begin{document}

\title{Motion Planning for Autonomous Driving:

The State of the Art and Future Perspectives}

\author{

Siyu Teng,  Xuemin Hu, Peng Deng, Bai Li, Yuchen Li, Yunfeng Ai, Dongsheng Yang, 

Lingxi Li,  Zhe Xuanyuan, Fenghua Zhu, ~\IEEEmembership{Senior Member,~IEEE,} Long Chen, ~\IEEEmembership{Senior Member,~IEEE,}

\thanks{This work was supported in part by National Natural Science Foundation of China under Grant 62273135 and 62103139; Natural Science Foundation of Hubei Province in China under Grant 2021CFB460; 2022 Opening Foundation of State Key Laboratory of Management and Control for Complex Systems under Grant E2S9021119; the Guangdong Provincial Key Laboratory of Interdisciplinary Research and Application for Data Science, BNU-HKBU United International College 2022B1212010006. Guangdong Higher Education Upgrading Plan with UIC research grant R0400001-22 and R201902. (Siyu Teng and Xuemin Hu contributed equally to this work). (Corresponding authors: Zhe Xuanyuan, Fenghua Zhu and Long Chen).

Siyu Teng and Yuchen Li are with BNU-HKBU United International College, Zhuhai, 519087, China and Hong Kong Baptist University, Kowloon, Hong Kong, 999077, China (e-mail: siyuteng@ieee.org).

Xuemin Hu and Peng Deng are with the School of Computer Science and Information Engineering, Hubei University, Wuhan 430062, China.

Bai Li is with the State Key Laboratory of Advanced Design and Manufacturing for Vehicle Body, and also with the College of Mechanical and Vehicle Engineering, Hunan University, Changsha 410082, China.


Yunfeng Ai is with University of Chinese Academy of Sciences, Beijing, 100049, China.

Dongsheng Yang is with the School of Public Management/Emergency Management, Jinan University, Guangzhou 510632, China.

Lingxi Li is with the Purdue School of Engineering and Technology, Indiana University-Purdue University Indianapolis (IUPUI), Indianapolis, USA.

Zhe Xuanyuan is with the Guangdong provincial key lab of IRADS, BNU-HKBU United International College, Zhuhai, 519087, China.

Fenghua Zhu and Long Chen are with Institute of Automation, Chinese Academy of Sciences,Beijing, China, 100190, and Long Chen is also with Waytous Ltd. (e-mail:fenghua.zhu@ia.ac.cn; long.chen@ia.ac.cn).

}
\thanks{Manuscript received April 19, 2021; revised August 16, 2021.}}

\markboth{IEEE TRANSACTIONS ON INTELLIGENT VEHICLES}%
{Shell \MakeLowercase{\textit{et al.}}: A Sample Article Using IEEEtran.cls for IEEE Journals and Transactions}


\maketitle

\begin{abstract}


Intelligent vehicles (IVs) have gained worldwide attention due to their increased convenience, safety advantages, and potential commercial value. Despite predictions of commercial deployment by 2025, implementation remains limited to small-scale validation, with precise tracking controllers and motion planners being essential prerequisites for IVs. This paper reviews state-of-the-art motion planning methods for IVs, including pipeline planning and end-to-end planning methods. The study examines the selection, expansion, and optimization operations in a pipeline method, while it investigates training approaches and validation scenarios for driving tasks in end-to-end methods. Experimental platforms are reviewed to assist readers in choosing suitable training and validation strategies. A side-by-side comparison of the methods is provided to highlight their strengths and limitations, aiding system-level design choices. Current challenges and future perspectives are also discussed in this survey.
\end{abstract}

\begin{figure} [t]
    \centering
	  \subfloat[Pipeline framework    \label{1a_123}]{
       \includegraphics[width=0.98\linewidth]{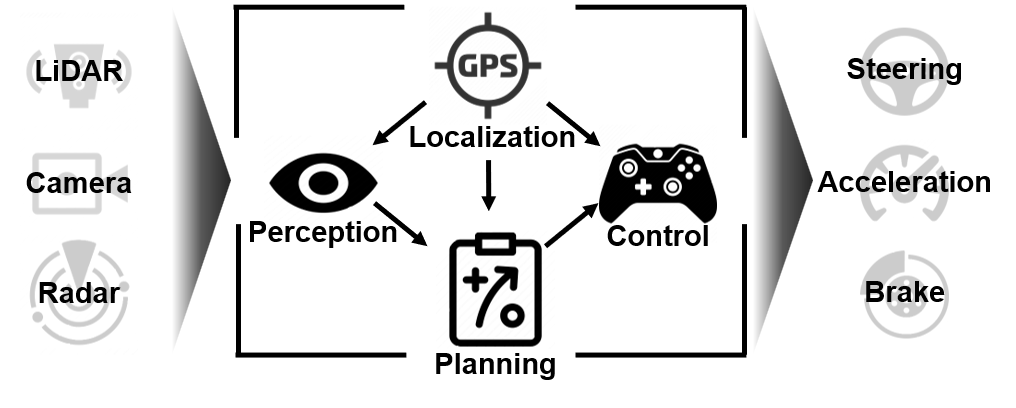}}

   \hfill

	  \subfloat[End-to-end framework \label{1b_123}]{
        \includegraphics[width=0.98\linewidth]{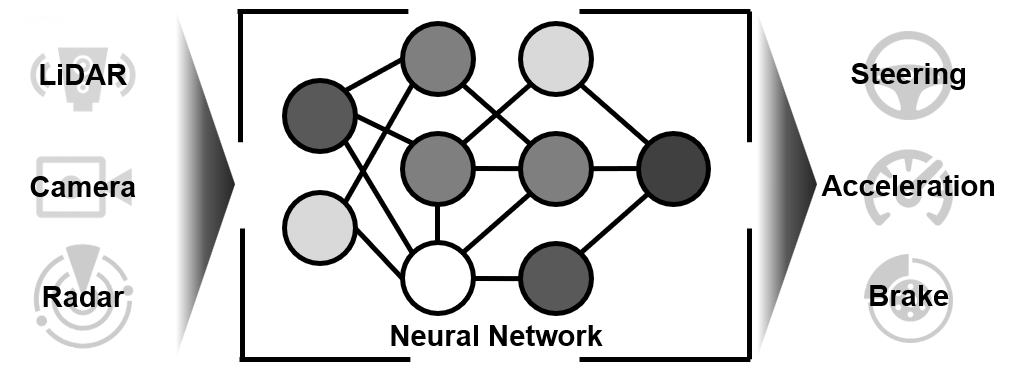}}
    \hfill
	  \caption{Pipeline and end-to-end frameworks surveyed in \cite{surveye2e}. The pipeline framework for autonomous driving consists of many interconnected modules, while the end-to-end method treats the entire framework as one learnable learning task.}
	  \label{fig:Total} 
\end{figure}

\section{Introduction}


\IEEEPARstart{I}{ntelligent} vehicles (IVs) have attracted significant interest from governments, industries, academia, and the public, owing to their potential to transform transportation through advances in artificial intelligence and computer hardware \cite{Surveyofsurvey}. The deployment of IVs holds great promise for reducing road accidents and alleviating traffic congestion, thereby improving mobility in densely populated urban areas \cite{wenshuowang}. Despite remarkable contributions by leading experts in the field, IVs remain primarily confined to limited trial programs due to concerns about their reliability and safety. To enhance situational awareness and improve safety, efficiency, and overall capabilities, IVs are equipped with a variety of sensors. However, even with an array of sensors, an IV stills face challenges in adequately detecting and responding to complex scenarios. Consequently, ensuring the safety, robustness, and adaptability of planning methods becomes crucial for the successful implementation of autonomous driving \cite{OS}.

\subsection{Background}

The pipeline planning method, also known as the rule-based planning method, is a well-established category of planners. As depicted in \ref{1a_123}, this method serves as a core component of the pipeline framework and must be integrated with other methods, such as perception \cite{TIV___1},  localization, and control, to accomplish autonomous driving tasks. A significant advantage of the pipeline framework is its interpretability, enabling the identification of defective modules when malfunctions or unexpected system behavior occur. In Section-\ref{ppl}, the focus is solely on the planning method within the pipeline framework. The pipeline planning method comprises two primary components: global route planning, which generates a road-level path from the origin to the destination, and local behavior/trajectory planning, which generates a short-term trajectory. Although widely used in the industry, the pipeline planning method requires substantial computational resources and numerous manual heuristic functions \cite{planning_k}. This study specifically addresses the expansion and optimization mechanisms of the pipeline planning method.

The end-to-end planning method, also known as the learning-based approach, is the sole component in the end-to-end framework and has become a trend in autonomous vehicle research. As illustrated in \ref{1b_123}, the entire driving framework is treated as a single machine learning task that converts raw perception data into control commands. The driving model acquires knowledge through imitation learning, develops driving policies through reinforcement learning, and continuously self-optimizes via parallel learning. Despite its appealing concept, determining the reasons for model misbehavior can be challenging. Consequently, this study focuses on the network structure, training techniques, and deployment tasks of the end-to-end model.

\subsection{Comparison}

\textcolor{black}{In this subsection, we provide a concise overview of the distinctions between pipeline and end-to-end methods, particularly highlighting their respective advantages and disadvantages.}

\textcolor{black}{The pipeline framework, widely implemented in the industry, allows engineers to focus on well-defined sub-tasks and independently improve each sub-model within the entire pipeline. Due to its clear intermediate representations and deterministic decision-making rules, this framework facilitates pinpointing the root cause of errors when unexpected behavior occurs. Moreover, it enables reliable reasoning about how the system generates specific control signals. However, the pipeline framework has some drawbacks. Individual sub-models may not be optimal for all driving scenarios, posing a challenge to the generalization of the framework. Additionally, the concatenation of sub-modules and the numerous manual customization constraints in each sub-model can compromise the robustness and real-time capabilities of the method.}

\textcolor{black}{The end-to-end framework optimizes the entire driving task, from raw perception to control signals, as a single deep learning task. By learning optimal intermediate representations for the target task, the framework can attend to any implicit sources of raw data without human-defined information bottlenecks, enhancing its generalization for various scenarios. The end-to-end framework's streamlined architecture, consisting of one or a few networks, also offers superior robustness and real-time capabilities compared to the pipeline framework. However, as research progresses, the end-to-end optimization faces a critical interpretability issue. Without intermediate outputs, tracing the initial cause of an error and explaining why the model arrived at specific control commands or trajectories becomes more challenging.}

\subsection{Paper Structure}

\textcolor{black}{In Section \ref{ppl}, pipeline planning methods are reviewed, including global route planning and local behavior/trajectory planning, with a particular focus on the expansion and optimization mechanisms. In Section \ref{e2e}, end-to-end planning methods are examined, encompassing imitation learning, reinforcement learning, and parallel learning, while exploring network architecture, generalizability, robustness, and validation \& verification methods. Additionally, large datasets, simulation platforms, and physical platforms play auxiliary roles in the development of autonomous driving with higher levels of intelligence and mobility. Therefore, other aspects of autonomous driving are summarized in Section \ref{experiments}, including datasets, simulation platforms, and physical platforms. Finally in Section \ref{future}, current challenges and future directions of autonomous driving are reviewed.}


\subsection{Contributions}

\textcolor{black}{This paper presents a comprehensive analysis of the general planning methods for autonomous driving. Broadly speaking, planning methods for autonomous driving can be classified into two categories: pipeline and end-to-end.} 

\textcolor{black}{There have been numerous state-of-the-art works on motion planning for IVs, however, a comprehensive review encompassing both pipeline and end-to-end methods has yet to be conducted. The pipeline is a classical planning method commonly used in the industry, with general categories outlined in previous research \cite{Li3, sur_pipeline2}. In this paper, we propose a new classification of pipeline methods that captures the extensively deployed approaches in a manner more relevant to industry selection, based on the expansion and optimization mechanisms of each method. Our proposed classification includes state grid identification, primitive generation, and other approaches. The end-to-end approach has emerged as a popular research direction in recent years, as demonstrated by previous work \cite{surveye2e, Review1}, which illustrates methods for mapping raw perception inputs to control command outputs. In this survey, we not only review the latest achievements in imitation learning (IL) and reinforcement learning (RL) but also introduce a novel category called parallel planning. This category proposes a virtual-real interaction confusion learning method for a reliable end-to-end planning method. Furthermore, we provide a thorough analysis and summary of the latest datasets, simulation platforms, and semi-open real-world testing scenarios, which serve as essential auxiliary elements for the advancement of IVs. To the best of our knowledge, this survey presents the first comprehensive analysis of motion planning methods in various scenarios and tasks.}

\section{Pipeline Planning Methods \label{ppl}}

The pipeline method, also known as the modular approach, is widely used in the industry and has become the conventional approach. This method originates from architectures that primarily evolved for autonomous mobile robots and consists of self-contained, interconnected modules such as perception, localization, planning, and control.

Planning methods are responsible for calculating a sequence of trajectory points for the ego vehicle's low-level controller to track, typically consisting of three functions: global route planning, local behavior planning, and local trajectory planning \cite{Li2,Li3}. \textcolor{black}{ Global route planning provides a road-level path from the start point to the end point on a global map. Local behavior planning decides on a driving action type (e.g., car-following, nudge, side pass, yield, and overtake) for the next several seconds. Local trajectory planning generates a short-term trajectory based on the decided behavior type. In fact, the boundary between local behavior planning and local trajectory planning is somewhat blurred \cite{Li2}, as some behavior planners do more than just identify the behavior type. For the sake of clarity, this paper does not strictly distinguish between the two functions, and the related methods are simply regarded as trajectory planning methods.}

\textcolor{black}{This section categorizes the related algorithms into two functions: global route planning and local behavior/trajectory planning. To provide a more detailed analysis and discussion, local behavior/trajectory planning is divided into three components: state grid identification, primitive generation, and other approaches, based on their respective extension methods and optimization theories}
\subsection{Global Route Planning}
Global route planning is responsible for finding the best road-level path in a road network, which is presented as a directed graph containing millions of edges and nodes. A route planner searches in the directed graph to find the minimal-cost sequence that links the starting and destination nodes. Herein, the cost is defined based on the query time, preprocessing complexity, memory occupancy, and solution robustness considered. Edsger Wybe Dijkstra is a pioneer in this field and innovatively proposes the Dijkstra algorithm \cite{GlobalPath_Dijkstra,Dij} named after him. Lotfi et al. \cite{Advance_dijkstra} construct a Dijkstra-based intelligent scheduling framework that computes the optimal scheduling for each agent, including maximum speed, minimum movement, and minimum consumption cost. A-star algorithm \cite{Astar,Astar69} is another famous algorithm in road-level navigation tasks, it leverages the advantages of the heuristic function to streamline research space. All of these algorithms substantially alleviate the problem of transportation efficiency and garnered significant attention in the field of intelligent transportation systems.

\subsection{Local Behavior/Trajectory Planning}
Local behavior planning and local trajectory planning functions work together to compute a safe, comfortable and continuous local trajectory based on the identified global route from route planning. 
Since the resultant trajectory is local, the two functions have to be implemented in a receding-horizon way unless the global destination is not far away \cite{Li6}. It deserves to emphasize that the output of the two functions should be a trajectory rather than a path \cite{Li7, tmach}, and the trajectory interacts with other dynamic traffic participants, otherwise, extra efforts are needed for the ego vehicle to evade the moving obstacles in the environment.

Nominally, local planning is done by solving an optimal control problem (OCP), which minimizes a predefined cost function with multiple types of hard or soft constraints satisfied \cite{SoftConstrain,Li16}. The solution to the OCP is presented as time-continuous control and state profiles, wherein the desired trajectory is reflected by a part of the state profiles. \textcolor{black}{As shown in Equ. \ref{equ:pipline}, the state space of the vehicle is denoted as $\bm{z} \in \mathbb{R}^{n_{z}}$, the control space is presented as $\bm{u} \in \mathbb{R}^{n_{u}}$. $\Upsilon$ shows the workspace. The obstacle space as $\Upsilon_{OBS}\subset \Upsilon$and the free space is described as  $\Upsilon_{FREE}\subset \Upsilon\setminus \Upsilon_{OBS}$.}

\begin{eqnarray}
    \begin{aligned}
        \min_{\bm{z}(t),\bm{u}(t),T}&{J (\bm{z}(t),\bm{u}(t))},\\
        s.t.,&\dot{\bm{z}}(t)=f(\bm{z}(t),\bm{u}(t));\\
        &\bm{z}\le \bm{z}(t)\le \bm{\bar{u}},\bm{u}\le \bm{u}(t)\le \bm{\bar{z}},,t\in [0,T];\\
        &\bm{z}(0)=\bm{z}_{init},\bm{u}(0)=\bm{u}_{init};\\
        &g_{end}(\bm{z}(T), \bm{z}(T)) \le 0;\\
        &fp(\bm{z}(t))\subset \Upsilon_{FREE},t\in [0,T];
    \end{aligned}
    \label{equ:pipline}
\end{eqnarray}

\textcolor{black}{
 The planning process duration in seconds is described as $T$, where $t\subset T$, and the cost function to be minimized is denoted as $J$. We use the common shorthand $\dot{\bm{z}}$ to denote the derivative with respect to time, $\dot{\bm{z}}=\partial \bm{z}/ \partial \bm{t}$, $\dot{\bm{u}}=\partial \bm{u}/ \partial \bm{t}$. The vehicle kinematic is described by the function $f$ and the allowable intervals where $\bm{z}(t)$ and $\bm{u}(t)$ are denoted by $[\bm{z},\bm{\bar{z}}]$ and $[\bm{u},\bm{\bar{u}}]$ respectively, where $\bm{z}$ and $\bm{u}$ representing the initial values. The inequality $g_{end} \le 0$ models the implicit end-point conditions at $t = T$. Finally, $f$ is a mapping from the vehicle state to its footprints, and $fp(\cdot ):\mathbb{R}^{n_{z}}\to \mathbb{R}^{2}$ represents the collision-avoidance constraints. In the following context of this section, we provide a detailed review of the motion planning method based on this scheme.
}

Since the analytical solution to such an OCP is generally not available, two types of operations are needed to construct a trajectory. Concretely, local planning is divided into three parts, the first type of operation is to identify a sequence of state grids, the second type is to generate primitives between adjacent state grids, and The third is an organic combination of the first two.

\subsubsection{State Grid Identification}  
State grid identification can be done by search, selection, optimization, or potential minimization. Search-based methods abstract the continuous state space related to the aforementioned OCP into a graph and find a link of states there. Prevalent search-based methods include A* search \cite{Li17} and dynamic programming (DP) \cite{Li19}.  Many advanced applications of these algorithms have pushed its influence to the top of the heap, such as Hybrid A*\cite{hybridAstar}, Bi-direction A*, Semi-optimization A*\cite{Bi-Astar}, and LQG framework \cite{Li19}. Selection-based methods decide the state grids in the next one or several steps by seeking the candidate with the optimal cost function. Greedy selection \cite{Li22} and Markov decision process (MDP) series methods typically \cite{Li23,Li24} fall into this category.

An optimization-based method discretizes the original OCP into a mathematical program (MP), the solution of which are high-resolution state grids \cite{Li28,Li29}. MP solvers are further classified as gradient-based and non-gradient-based ones; gradient-based solvers typically solve nonlinear programs \cite{Li16}, quadratic programs \cite{Li21,Li28,Li31}, quadratically constrained quadratic programs \cite{Li30} and mix-integer programs; non-gradient-based solvers are typically represented by metaheuristics \cite{Li35}. Multiple previous methods could be combined to provide a coarse-to-fine local behavior/motion planning strategy.
\begin{figure*}[t]
    \centering
    \includegraphics[width=0.88\linewidth]{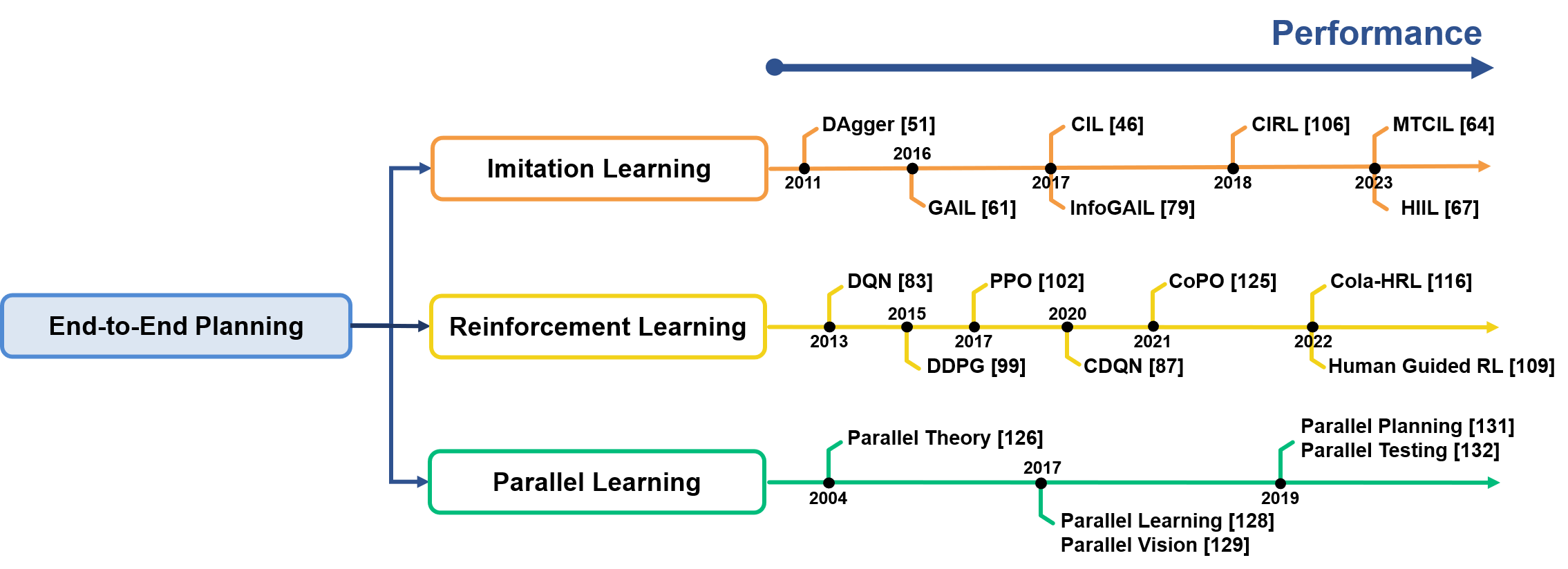}
    \caption{The critical method survived in End-to-End Planning Section. The time axis (dark blue) represents the progressiveness of the survived methods, and the performance of the methods is better with the latter proposed time.}
    \label{fig:Mindmap}
\end{figure*}
\subsubsection{Primitive Generation} 
Primitive generation commonly manifests as closed-form rules, simulation, interpolation, and optimization. Closed-form rules stand for methods that compute primitives by analytical methods with closed-form solutions. Typical methods include the Dubins/Reeds-Shepp curves \cite{Li42}, polynomials \cite{Li22}, and theoretical optimal control methods \cite{Li44, hu2018dynamic}. Simulation-based methods generate trajectory/path primitives by forwarding simulation, which runs fast because it has no degree of freedom \cite{Li17}. Interpolation-based methods are represented by splines or parameterized polynomials \cite{Li46}. Optimization-based methods solve a small-scale OCP numerically to connect two state grids \cite{Li47,Li48}.

\subsubsection{Other Approaches}
State grid identification and primitive generation are two fundamental operations to construct a trajectory. Both operations may be organized in various ways. For example, Kuwata et al. \cite{Li45} integrate both operations in an iterative loop; Hu et al. \cite{Li47} build a graph of primitives offline before online state grid identification; Fan et al. \cite{Li21} identify the state grids before generating connective primitives. If a planner only finds a path rather than a trajectory, then a time course should be attached to the planned path as a post-processing step \cite{Li48}. This strategy, denoted as path velocity decomposition (PVD), has been commonly used because it converts a 3D problem into two 2D ones, which largely facilitates the solution process. Conversely, non-PVD methods directly plan trajectories, which has the underlying merit to improve the solution optimality \cite{Li19,Li20,Li51,li12342}.

Recent studies in this research domain include how to develop specific planners that fit specific scenarios/tasks particularly \cite{Li6,Li51}, and how to plan safe trajectories with imperfect upstream/downstream modules \cite{Li51}. The past decades have seen increasingly rapid progress in the autonomous driving field. In addition to the advances in computing hardware, this rapid progress has been enabled by major theoretical progress in the computational aspects of mobile robot motion planning theory. Research efforts have undoubtedly been spurred by the improved utilization and safety of road networks that intelligent vehicles would provide.

\section{End-to-End Planning Methods \label{e2e}}

\textcolor{black}{End-to-end stands for the direct mapping from raw sensor data into trajectory points or control signals. Because of its ability to extract task-specific policies, it has achieved great success in a variety of fields \cite{Review1}.} Compared with the pipeline method, there is no external gap between the perception and control modules, and seldom human-customized heuristics are embedded, so the end-to-end method deals with vehicle-environment interactions more efficiently. End-to-end has a higher ceiling, with the potential to achieve expert performance in the autonomous driving field. \textcolor{black}{This section categorizes the end-to-end method into three distinct types from learning methods: imitation learning using supervised learning, reinforcement learning utilizing unsupervised learning, and parallel learning incorporating confusion learning. Fig. \ref{fig:Mindmap} further clarification of the structural relationships fo end-to-end planner, highlighting the performance and progressiveness of the reviewed methods.}


\subsection{Imitation Learning}
Imitation learning (IL) refers to the agent learning policy based on expert trajectory, which generally provides expert decisions and control information \cite{TIV221}. Each expert trajectory contains a sequence of states and actions, and all ``state-action" pairs are extracted to construct datasets. In the IL task, the model leverages constructed dataset to learn the latent relationship between state and action, the state stands for a feature and the action demonstrates labels. Thus, the specific objective of IL is to appraise the most fitness mapping between state and action, so that the agent achieves the expert trajectories as much as possible. \textcolor{black}{The formulation of IL is summarized in summarized as follows:}

\textcolor{black}{
Given a dataset $\mathbb{D}$ comprising ``state-action" pairs $(s, a)$ generated from expert trajectories $\pi^{*}$, the primary objective of IL is to learn a policy $\pi_{\theta }(s)$ that closely approximates the expert trajectories in any input state $s$, as determined by Equ. \ref{equ:IL}:}

\textcolor{black}{
\begin{equation}
    \underset{\theta } {{\arg \min}} \, \mathbb{E}_{s\sim P(s|\theta)}L(\pi^{*}(s), \pi_{\theta}(s)),
    \label{equ:IL}
\end{equation}}

\noindent\textcolor{black}{where $P(s|\theta)$ stands for the distribution of the current state from the trained policy $\pi_{\theta}$.}

\begin{table*}
\centering
\caption{THE CRUCIAL REVIEWS AND RELATIVE INFORMATION OF EACH FAMOUS END-TO-END MODELS IN AUTONOMOUS DRIVING.}
\label{tab:imitationlearning}

\begin{tabular}{m{2.25cm} c m{2cm}<{\centering}  m{2.5cm}<{\centering}  m{2.4cm}<{\centering} m{3.5cm} m{2cm}<{\centering}}
\hline
\rowcolor[rgb]{0.937,0.937,0.937} \multicolumn{1}{c}{\textbf{Article}} & \textbf{Category} & \multicolumn{1}{c}{\textbf{Input}}      & \multicolumn{1}{c}{\textbf{Output}} & \multicolumn{1}{c}{\textbf{Implement Tasks}} & \multicolumn{1}{c}{\textbf{Auxiliary Method}}                                              & \multicolumn{1}{c}{\textbf{\textbf{Dataset}}}        \\ 
\hline
Bojarski et al.\cite{IL2}                                                        & \textbf{BC}       & monocular image                         & steering angle                      & lane Keeping                                 & CNN is the only component of end-to-end model                                              & physical $\&$ simulate platform    \\
Codevilla  et al.\cite{CIL}                                                       & \textbf{BC}       & monocular image                         & control information                 & simulation navigation task                   & High-level commands as a switch to select the branch                                       & Carla                                                \\
Chen et  al.\cite{IL_Inter1}                                                           & \textbf{BC}       & monocular image                         & control information                 & simulation navigation task                   & Affordance is used to predict control actions                                              & TORCS Dataset $\&$ KITTI                              \\
Sauer et  al.\cite{IL_Inter2}                                                          & \textbf{BC}       & monocular video $\&$ directional input   & control information                 & physical navigation task                     & Conditional affordance is trained to calculate intermediate representations                & Carla                                                \\
Zeng et al.\cite{IL_Inter4}                                                            & \textbf{BC}       & Lidar data $\&$ HD Map                   & trajectory, scenario representation & physical navigation task                     & The intermediate representation is used to improve the model's interpretability            & physical dataset collected in North America          \\
Sadat  et al.\cite{IL_Inter5}                                                          & \textbf{BC}       & Lidar data $\&$ HD Map                   & trajectory, scenario representation & physical navigation task                     & A joint system with~interpretable intermediate representations for E2E planner   & physical dataset collected in North America          \\
Ross et  al.\cite{IL8}                                                            & \textbf{DPL}      & monocular image                         & control information                 & autonomous racing competition                & An iterative algorithm is proposed to guarantee the performance in corner cases            & 3D racing simulator                                     \\
Zhang et  al.\cite{IL10}                                                          & \textbf{DPL}      & monocular image                         & control information                 & autonomous racing competition                & Embedded query-efficient model to reduce the request for expert trajectories           & racing car simulator               \\
Yan et  al.\cite{IL12}                                                            & \textbf{DPL}      & LiDAR, ego-vehicle speed, Sub-goal      & control information                 & physical $\&$ simulation navigation task      & The novice and the expert policy is fused to control the robot                             & physical and simulate platform    \\
Li et  al.\cite{OIL}                                                              & \textbf{DPL}      & monocular image $\&$ sub-goal            & waypoint, control information       & autonomous racing                            & A reward-based online method learns from multiple experts                                  & Sim4CV                                               \\
Ohn-Bar  et al.\cite{LSD}                                                        & \textbf{IRL}      & monocular image                         & control information                 & simulation navigation task                   & Scenario context is embedded into the policy learning network                              & Carla                                                \\
Levine  et al.\cite{BIRL2}                                                         & \textbf{IRL}      & BEV image    & control information                 & keep the lane, change lanes $\&$ takeover     & The Gaussian algorithm is used to learn the relevance of features in expert trajectories.  & Highway driving simulator                            \\
Brown  et al.\cite{BIRL3}                                                          & \textbf{IRL}      & monocular image                         & control information                 & keep the lane, change lanes $\&$ takeover     & The high-confidence upper bounds on the $alpha$-worst-case are embedded into the policy network. & Highway driving simulator                            \\
Palan et  al.\cite{BIRL4}                                                          & \textbf{IRL}      & monocular image                         & control information                 & keep the lane, change lanes $\&$ takeover     & A globally normalized reward function is constructed.                                      & Lunar lander simulator                               \\
Ziebart  et al.\cite{MEIRL}                                                        & \textbf{IRL}      & Road network, Sub-goal, $\&$ GPS Data    & control information                 & long range autonomous navigation task        & A probabilistic approach is proposed for maximum entropy                                   & Driver route modeling                                \\
Lee et  al.\cite{MEIRL4}                                                            & \textbf{IRL}      & monocular image                         & control information, costmap        & keep the lane, change lanes $\&$ takeover     & The query generation process is used to improve the generalization                         & NGSIM $\&$ Carla                                      \\
Ho et  al.\cite{MEIRL5}                                                             & \textbf{IRL}      & monocular image                         & control information                 & keep the lane, change lanes $\&$ takeover     & GAN is integrated into the end-to-end model                                                & Carla                                                \\
Phan  et al.\cite{Extend}                                                           & \textbf{IRL}      & BEV image, HD map, obstacle information & Control information                 & physical navigation task                      & A three-step IRL planner is proposed                                                       & physical dataset from the Las Vegas Strip  \\
\hline
\end{tabular}
\end{table*}

\textcolor{black}{
Based on this formulation, three widely used training methods are survived in this part \cite{IL1}, first manifests as a negative method, named behavioral cloning (BC); The second builds on BC, named direct policy learning (DPL); The last is a task-dependent method, named inverse reinforcement learning (IRL) method.  Table \ref{tab:imitationlearning} presents all famous imitation learning methods reviewed in this part.}
\subsubsection{Behavioral Cloning}
Behavioral Cloning (BC) manifests as the primary method of IL in autonomous driving \cite{IL2,zhu2022multi}. \textcolor{black}{The agent leverages expert trajectories to the training model and then replicates the policy using a classifier/regressor. BC is a passive method, where the objective is to learn the target policy by passively observing the complete execution of commands. This requires the premise that the state-action pairs in all trajectories are independent.}

\begin{figure}
    \centering
    \includegraphics[width=0.98\linewidth]{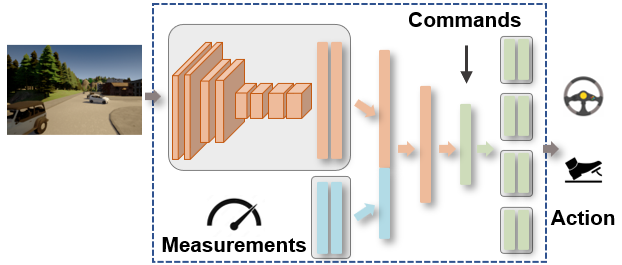}
    \caption{The model proposed in \cite{CIL}. Measurements stand for the velocity of ego-vehicle. The high-level command, including straight, left, right, and lane following. Actions are control signals including steering, accretion, and brake.}
    \label{fig:cil}
\end{figure}

Bojarski et al. \cite{IL2} construct a pioneering framework for BC, which trains a convolutional neural network to only compute steering from a front-view monocular camera. This method exclusively outputs lateral control while ignoring longitudinal commands, rendering it can only be implemented in a limited number of uncomplicated scenarios.
Codevilla et al. \cite{CIL} proposed a famous IL model, named conditional imitation learning (CIL), which contains both lateral and longitudinal control, as shown in Fig. \ref{fig:cil}. Monocular images, velocity measurement of ego-vehicle, and high-level commands (straight, left, right and lane following) are used as input to CIL, and both predicted longitude and latitude control commands as output. Each command acts as a switch to select a specialized sub-module. CIL is a milestone for the CL method in autonomous driving and demonstrates that the  convolutional neural network (CNN) can learn to perform lane and road tracking tasks autonomously.

Based on CIL \cite{CIL}, many researchers include additional information such as global route, location information, or point cloud in the input stage \cite{IL6,IL7,T1}. These methods demonstrate strong generalization and robustness in various conditions, because of sufficient perception data input.

Because of its novel structure, IL methods exclude uncertainty estimation among different sub-systems and lead to fewer feedback milliseconds. However, this characteristic  leads to a significant drawback, lack of interpretability, which does not provide sufficient reasons to explain the decisions. Many researchers try to address this pain point by inserting the intermediate representation layer. Chen et al. \cite{IL_Inter1} propose a novel paradigm, named direct perception method, to predict an affordance for urban autonomous driving scenarios. The affordance represents a BEV format that clearly displays features about the surrounding environment and then is fed to a low-level controller to generate steering and acceleration. Sauer et al. \cite{IL_Inter2} further propose an advanced direct perception model, which leverages video and high-level commands  to intermediate representations and computes control signals as output. Compared with \cite{IL_Inter1}, this model can handle complex scenarios in urban traffic scenarios.
\textcolor{black}{Urtasun and her team also propose two interpretable end-to-end planners \cite{IL_Inter4, IL_Inter5}, which leverage raw LiDAR data and a High-Definition Map (HD Map) to predict safe trajectories and intermediate representations. The representations demonstrate how the policy responds to surrounding scenarios. Compared with previous methods that only use the monocular images as inputs, these planners can predict trajectories more safely,  owing to the resource-rich input information.}



The main feature of the BC method is that only experts can generate training examples, which directly leads to the training set being a subset of the states accessed during the execution of the learned policy \cite{hu2020learning}. Therefore, when the dataset is biased or overfitted, the method is limited to generalize. Moreover, when the agent is guided to an unknown state, it is hard to learn the correct recovery behavior.

\subsubsection{Direct Policy Learning} Direct Policy Learning (DPL), a training method based on BC, evaluates the current policy and then obtains more suitable training data for self-optimization. Compared with BC, the main advantage of DPL leverages expert trajectories to instruct the agent how to recover from current errors \cite{IL1}. In this way, DPL alleviates the limitation of BC due to insufficient data. In this section, we summarize a series of DPL methods.

\begin{figure} [t]
    \centering
	  \subfloat[DAgger]{
       \includegraphics[width=0.48\linewidth]{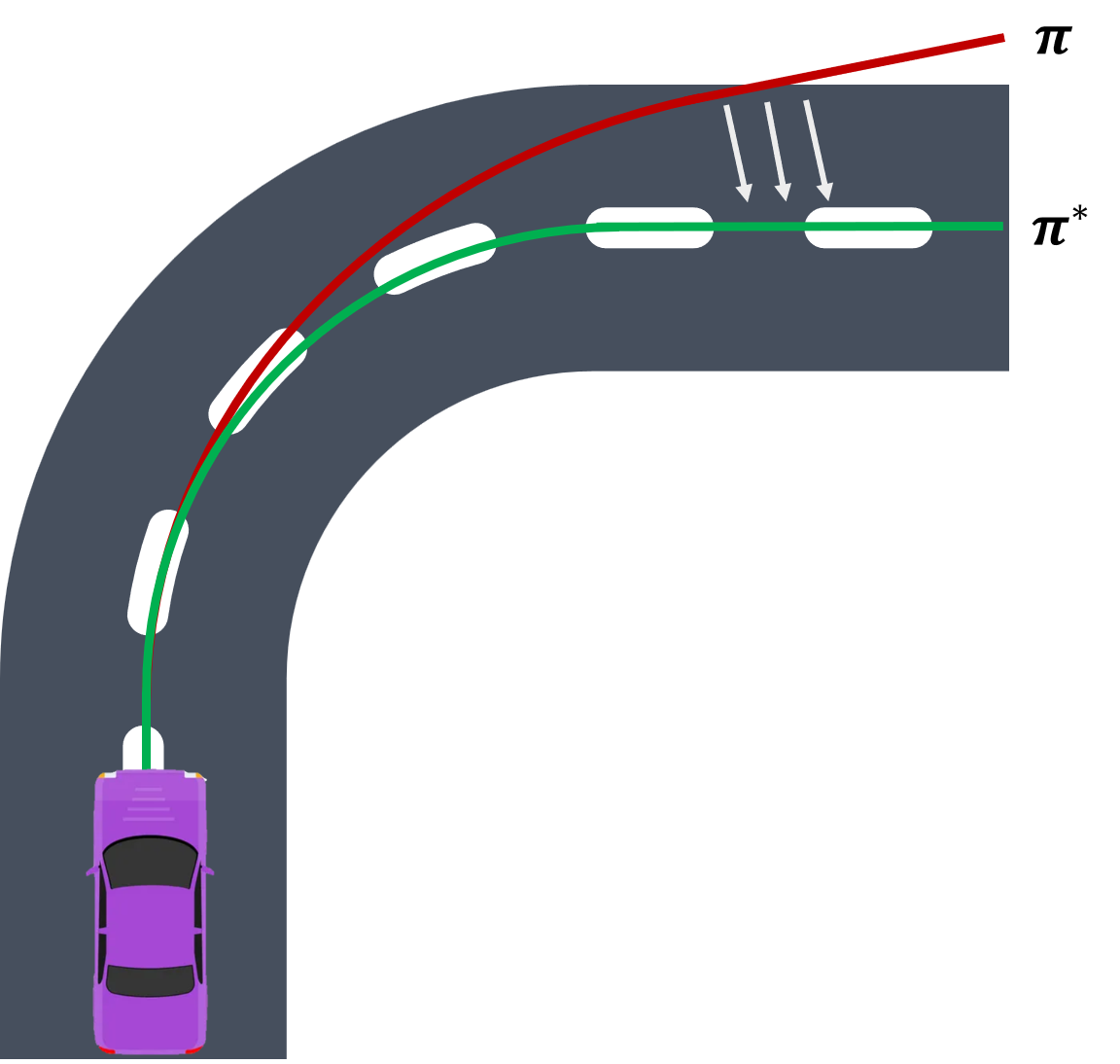}}
    \label{1a}\hfill
	  \subfloat[DAgger by coaching]{
        \includegraphics[width=0.48\linewidth]{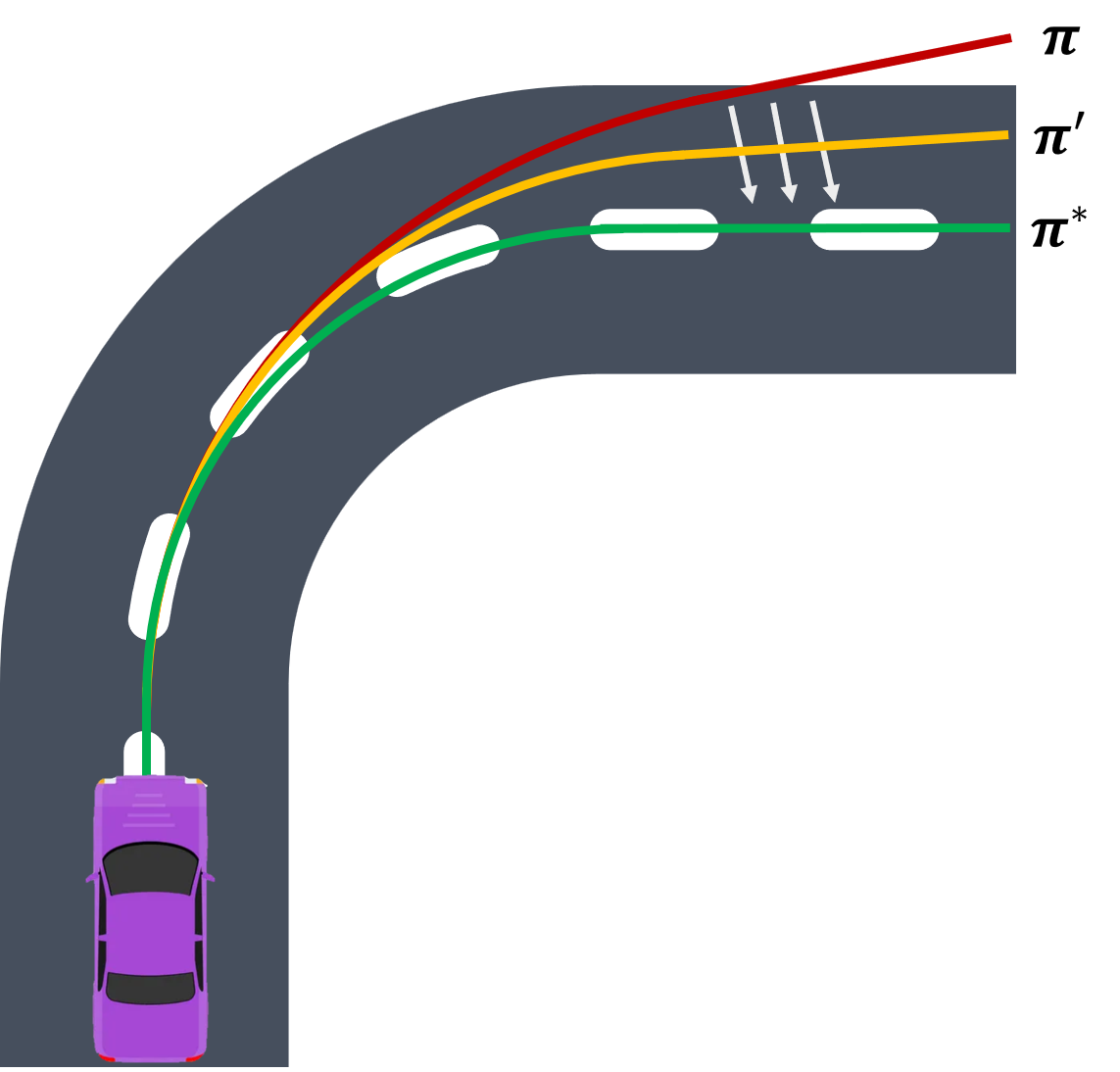}}
    \label{1b}\hfill
	  \caption{The DAgger method \cite{IL8} for Autonomous Driving Navigation Task.}
	  \label{fig:DAgger} 
\end{figure}

Ross et al. \cite{IL8} construct a classical online IL method named Dataset Aggregation (DAgger) method. This is an active method based on the Follow-the-Leader algorithm \cite{IL1}, each validation iteration is an online learning example. The method modifies the main classifier or regressor on all state$-$action pairs experienced by the agent. DAgger is a novel solution for sequential prediction problems, however, its learning efficiency might be suppressed by the far distance between policy space and learning space. In reply, He et al. \cite{IL9} propose a DAgger by coaching algorithm which employs a coach to demonstrate easy-to-learn policies for the learner and the demonstrated policies gradually converge to label. To better instruct the agent, the coach establishes a compromised policy which is not much worse than a ground truth control signal and much better than novice predicted action. As shown in Fig. \ref{fig:DAgger}, $\pi$ is the predicted command, $\pi^{*}$ shows the expert trajectory, and $\pi^{'}$ presents the compromised trajectory. $\pi^{'}$ is much easier than $\pi^{*}$ for agent to learn sub-optimal policy in each iteration, and the policy is asymptotically optimal.


Other researchers also point out some drawbacks of DAgger methods \cite{IL8,IL9}: inefficient query, inaccurate data collector, and poor generalization. In reply, Zhang et al. \cite{IL10} propose the SafeDAgger algorithm, which intends to improve the query efficiency of DAgger and can further reduce the dependence on label accuracy. Hoque et al. \cite{IL11} propose a ThriftyDAgger model, which integrates human feedback on corner cases, Yan et al. \cite{IL12} propose a novel DPL training scheme for navigation tasks in mapless scenarios, both of them improve the generalization and robustness of the model.

DAgger-based methods reduce dataset dependency and improve learning efficiency, however, these methods cannot distinguish between good or bad expert trajectories and ignore the learning opportunity from unfitness behaviors. In reply, Li et al. \cite{OIL} propose the observational imitation learning (OIL) method, which predicts the control commands from the monocular image and embeds waypoints as intermediate representations. OIL manifests as an online learning policy based on a reward function, it could learn from multi-experts and abandon the wrong policies.

\begin{figure} [b]
    \centering
    \includegraphics[width=0.9\linewidth]{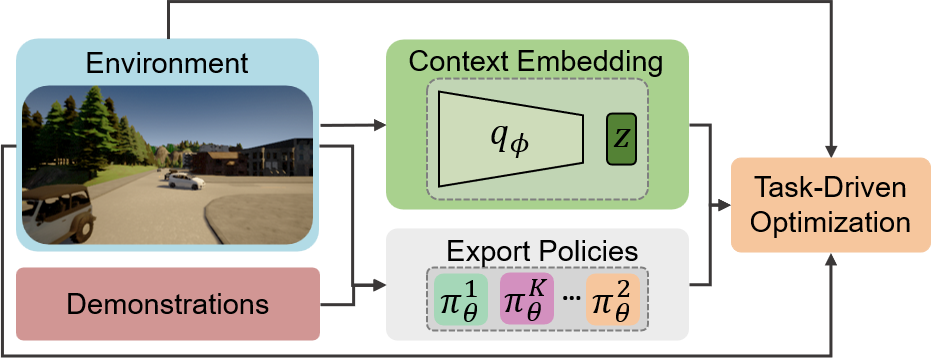}
    \caption{Training policies propose in \cite{LSD}. Export policies learn sub-optimal policy. Context Embedding is trained to learn scenarios. Both Context Embedding and Export Policies are fine-tuning in Task-Driven Optimization by Online method.}
    \label{fig:LSD}
\end{figure}

To fine-tune the agent policy in perception-to-action methods, Ohn-Bar et al. \cite{LSD} propose a method for optimizing situational driving policies which effectively captures reasoning in different scenarios, shown in Fig. \ref{fig:LSD}. The training policy is divided into three parts. First, the model learns sub-optimal policies by the BC method. Second, context embedding is trained to learn scenario features. Third, refined the integrated model by online interaction with the simulation and collect better data by a DAgger-based method. 

DPL is an iterative online learning policy that alleviates the requirements for the volume and distribution of dataset, while facilitating the continuous improvement of policies by effectively eliminating incorrect ones.
\subsubsection{Inverse Reinforcement Learning}
Inverse reinforcement learning (IRL) is designed to circumvent the drawbacks of the aforementioned methods by inferring the latent reasons between input and output. Similar to the prior methods, IRL needs to collect a set of expert trajectories at the beginning. However, instead of simply learning a state-action mapping, these expert trajectories are first inferred and then the behavioral policy is optimized based on the elaborate reward function.  IRL method can be classified into three distinct categories, max-margin methods, Bayesian methods, and maximum entropy methods.


The max-margin method leverages expert trajectories to evaluate a reward function that maximizes the margin between the optimal policy and estimates sub-optimal policies. these methods represent reward functions with a group of features utilizing a linear combination algorithm, where all features are considered independent.


Andrew Wu \cite{IRL1} is a pioneer in this field, he introduces the first max-margin IRL method, which puts forward three algorithms for computing the refined reward function. Furthermore, Pieter et al. \cite{IRL2} devise an optimized algorithm based on \cite{IRL1}, which assumes that an expert reward function can be expressed as a manually crafted linear combination of known features, with the objective of uncovering the latent relationships between weights and features.


The limitation of prior methods is that the quality and distribution of the expert trajectories sets an upper bound on the performance of the method. In reply, Umar et al. \cite{IRL3} propose a game-theoretic-based IRL method named multiplicative weights for apprenticeship learning, it has the capability to import prior policy to the agent about the weight of each feature and leverages a linear programming algorithm to modify the reward function so that its policy is stationary.

\begin{figure}
    \centering
    \includegraphics[width=0.9\linewidth]{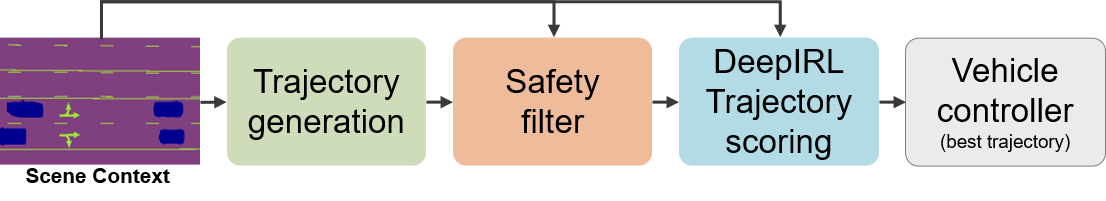}
    \caption{The system proposed in \cite{Extend} is divided into three stages: trajectory generation, safety filtering, and trajectory scoring.}
    \label{fig:threepart}
\end{figure}

In addition, Phan-Minh et al. \cite{Extend} propose an interpretable planning system, as shown in Fig. \ref{fig:threepart}. The trajectory generation module leverages perception information to compute a set of future trajectories. The safety filter is used to guarantee the basic safety with an interpretable method. DeepIRL trajectory scoring the predicted trajectories, which is the core contribution of this system. Furthermore, \cite{IRL4} and \cite{IRL5} propose preference-inference formulation, users can choose actions according to their personal preferences, which indeed improves the performance of the model.

The second part of IRL is Bayesian methods, which often leverage the optimized trajectory or the prior distribution of the reward to maximize the posterior distribution of the reward. 
The first Bayesian IRL is proposed by Ramachandran et al. \cite{BIRL1}, which references the IRL model from a Bayesian perspective and inferences a posterior distribution of the estimated reward function from a prior distribution. Levine et al. \cite{BIRL2} integrate a kernel function into the Bayesian IRL model \cite{BIRL1} to improve the accuracy of estimating reward and promote the performance in unseen driving. 

Furthermore, Brown et al. \cite{BIRL3} construct a sampling-based Bayesian IRL model, which utilizes expert trajectories to calculate practical high-confidence upper bounds on the $\alpha$-worst-case difference in expected return under the unseen scenarios without a reward function.
Palan et al. \cite{BIRL4} propose DemPref model, which utilizes the expert trajectory to learn a coarse reward function, the trajectory is used to ground the (active) query generation process, to improve the quality of the generated queries. DemPref alleviates the efficiency problems faced by standard preference-based learning methods and does not exclusively depend on high-quality expert trajectories.
 

The third part of IRL is the maximum entropy method, which is defined by using maximum entropy in the optimization routine to estimate the reward function. Compared with the previous IRL method, Maximum entropy methods are preferable for continuous spaces and have the potential ability to address the sub-optimal impact of expert trajectories. The first Maximum Entropy IRL model is proposed by Ziebart \cite{MEIRL}, which leverages the same method as \cite{IRL1} and could alleviate both noises and imperfect behavior in the expert trajectory. The agent attempts to optimize the reward function under supervision by linearly mapping features to rewards.

And then, many researchers \cite{MEIRL3,MEIRL4,MEIRL5} implement the maximum entropy IRL to physical end-to-end autonomous driving. Among them, \cite{MEIRL5} propose Generative Adversarial Imitation Learning (GAIL), which has become a classical algorithm in this field. GAIL leverages a generative adversarial network (GAN) to generate the distribution of expert trajectories with a model-free method in order to alleviate the problem of state drift caused by insufficient datasets. Because of sufficient reconstruction expert trajectories and competitive policies, GAIL achieves performance comparable to that of human drivers in specific scenarios. Based on \cite{MEIRL5}, many works have been proposed, such as InfoGAIL \cite{GIAL1}, Directed-InfoGAIL \cite{GAIL3}, Co-GAIL \cite{GIAL2}, all of them achieve competitive results in their implement fields. 

IRL provides several excellent works for autonomous driving, however, like the aforementioned methods, it also has long tail problems in corner cases. How to effectively improve the robustness and interpretability of IRL is also a future direction.

\textcolor{black}{The objective of IL methods is to acquire state-to-action mapping from expert trajectories. However, the generalizability of the method may be compromised when the dataset is intrinsically flawed (e.g., overfitting or uneven distribution) \cite{hu2020learning}. Additionally, when the agent is guided to an unknown state, predicting the correct behavior becomes a formidable challenge. To overcome these limitations, many researchers \cite{CIL, T1, zhu2022multi} have significantly enriched the dataset distribution using data augmentation and the combination of real and virtual data. These efforts ensure the generalizability of the methods and obtain competitive results.}

\subsection{Reinforcement Learning}

IL methods require large amounts of manually labeled data, and diverse drivers may arrive at entirely distinct decisions when presented with identical situations, which leads to uncertainty quandaries during training. In order to obviate the hunger for labeled data, some researchers have endeavored to utilize reinforcement learning (RL) algorithms for autonomous decision planning. \textcolor{black}{Reinforcement learning refers to the agent learning policy by interacting with an environment. Rather than imitating expert behavior, the goal of an RL agent is to maximize the cumulative numerical rewards from its environment via trial and error.} By consistently interacting with the environment, the agent gradually acquires knowledge of the optimal policy to attain the target. 

\textcolor{black}{Markov decision processes (MDPs) are typically used to formulate the RL problem. An MDP consists of a state space $\mathcal{S}$, and an action space $\mathcal{A}$, a reward function $R:\mathcal{S} \times \mathcal{A} \rightarrow \mathbb{R}$, a transition function $T:\mathcal{S} \times \mathcal{A} \times \mathcal{S} \rightarrow [0,1]$, and a discount factor $\gamma$ that trades instantaneous over future rewards, i.e. a tuple $(\mathcal{S,A},R,T,\gamma)$. At each time step $t$, the agent finds itself in a state $s \in \mathcal{S}$ and selects an action $a \in \mathcal{A}$ according to its policy $\pi: \mathcal{S} \times \mathcal{A} \rightarrow [0,1]$. Then the environment enters a new state $s^\prime \in \mathcal{S}$ with a transition probability $T$, where a new action is selected, and so on. Along with the state transition, the environment also gives rise to rewards $r$, special numerical values that the agent seeks to maximize over time through its choice of actions.
The goal is to find the optimal policy $\pi^*$, which results in the highest expected sum of discounted rewards \cite{sutton_book}:}

\textcolor{black}{
\begin{equation}
    \pi^* = \underset{\pi}  {{\arg\max}} \,  \mathbb{E}_{\pi} \left [ \sum_{t=0}^{N-1} \gamma^t r_{t+1} \mid s_0 = s \right ], 
\end{equation}
}

\noindent\textcolor{black}{where the initial states $s \in \mathcal{S}$. The horizon $N$ is the number of time steps, and reward $r_t=R(s_t,a_t)$, and $\gamma \in [0, 1]$ is the discount factor. The expectation means that the agent takes actions following the policy $\pi$ and gets corresponding discounted total rewards.}


\textcolor{black}{Based on this formulation, two main RL approaches to achieve optimal policies are developed, e.g., value-based reinforcement learning and policy-based reinforcement learning. Furthermore, based on those approaches, hierarchical reinforcement learning (HRL) and multi-agent reinforcement learning (MARL) are promising ways to solve more complex problems and better fit real driving scenarios. Training autonomous vehicles with RL methods has become a growing trend in end-to-end autonomous driving research.}



\subsubsection{Value-based Reinforcement Learning}

Value-based methods try to estimate the value of different actions in a given state and learn to assign a value to each action based on the expected reward that can be obtained by taking that action in that state. The agent learns to associate the rewards with the states and actions taken in the environment and leverages this information to make optimal decisions \cite{JAS_planning}.

\begin{figure}
    \centering
    \includegraphics[width=0.98\linewidth]{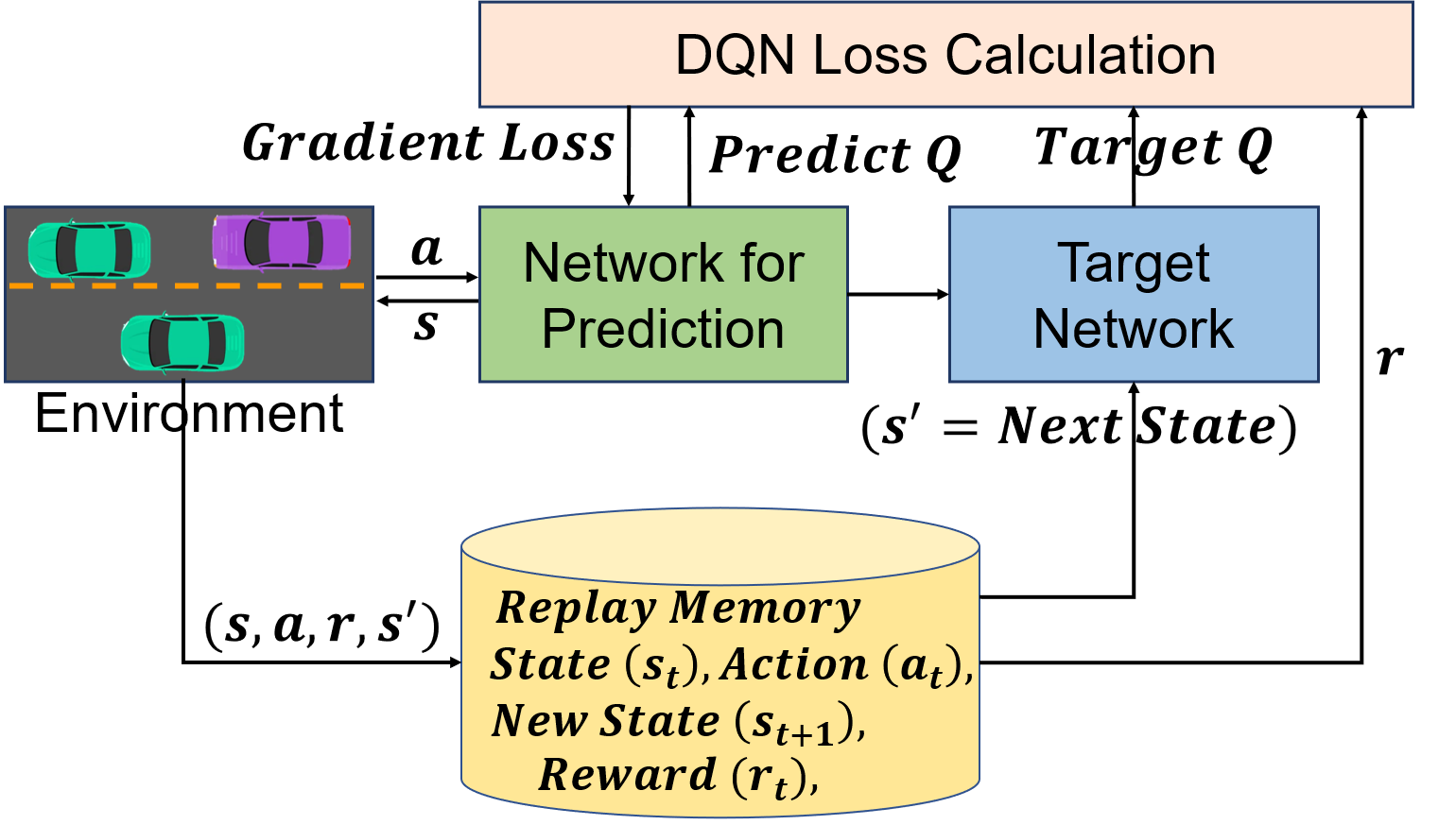}
    \caption{The architecture of DQN-based end-to-end  autonomous driving method.}
    \label{fig:DQN}
\end{figure}

Among value-based methods, Q-Learning \cite{Qlearning} stands out as the most prominent. The framework for implementing Q-Learning in end-to-end planning is illustrated in Fig. \ref{fig:DQN}. 
Mnih et al. \cite{VEL1} propose the first deep learning method by a Q-learning based approach that learns directly from screenshots to control signals. Furthermore, Wolf et al. \cite{RL-wolf2017} introduce the Q-learning method into the intelligent vehicle field, they define five different driving maneuvers in the Gazebo simulator \cite{RL-Gazebo}, and the vehicle chooses a corresponding maneuver based on the image information. For the purpose of alleviating the problem of poor stability  with high-dimensional perception input. The Conditional DQN \cite{ chen2020conditional} method is proposed, which leverages a defuzzification algorithm to enhance the predictive stability of distinct motion commands. The proposed model achieves a performance comparable to human driving in specific scenarios

In order to perform high-level decision-making for IVs on specific scenarios, Alizadeh et al. \cite{Alizadeh2019} train a DQN agent combined with DNN which outputs two discrete actions. The safety and agility of the ego vehicle can be balanced on-the-go, indicating that the RL agent can learn an adaptive behavior. Furthermore, Ronecker et al. \cite{Ronecker2019} propose a safer navigating method for IVs in highway scenarios by combining Deep Q-Networks from control theory. The proposed network is trained in simulation for central decision-making by proposing targets for a trajectory planner, which shows that the value-based RL can produce efficient and safe driving behavior in highway traffic scenarios.

The security of end-to-end autonomous driving also raises significant apprehension. Constrained Policy Optimization (CPO) \cite{SafeRL1} is a pioneering general-purpose policy exploit algorithm for constrained reinforcement learning with guarantees for near-constraint satisfaction at each iteration. Building on this, \cite{SafeRL2} and \cite{SafeRL3} present the Safety Gym benchmark suite and validate several constrained deep RL algorithms under constrained conditions. Li et al. \cite{RLTsy7} introduce a risk awareness algorithm into DRL frameworks to learn a risk-aware driving decision policy for lane-changing tasks with the minimum expected risk. Chow et al. \cite{SafeRL4} propose safe policy optimization algorithms that employ a Lyapunov-based approach \cite{Lya} to address CMDP problems. Furthermore, Yang et al. \cite{SafeRL5} construct a model-free safe RL algorithm that integrates policy and neural barrier certificate learning in a stepwise state constraint scenario.  Mo et al. \cite{RLTsy9} leverage Monte Carlo Tree Search to reduce unsafe behaviors on overtaking subtasks at highway scenarios.


\begin{figure}[b]
    \centering
    \includegraphics[width=0.98\linewidth]{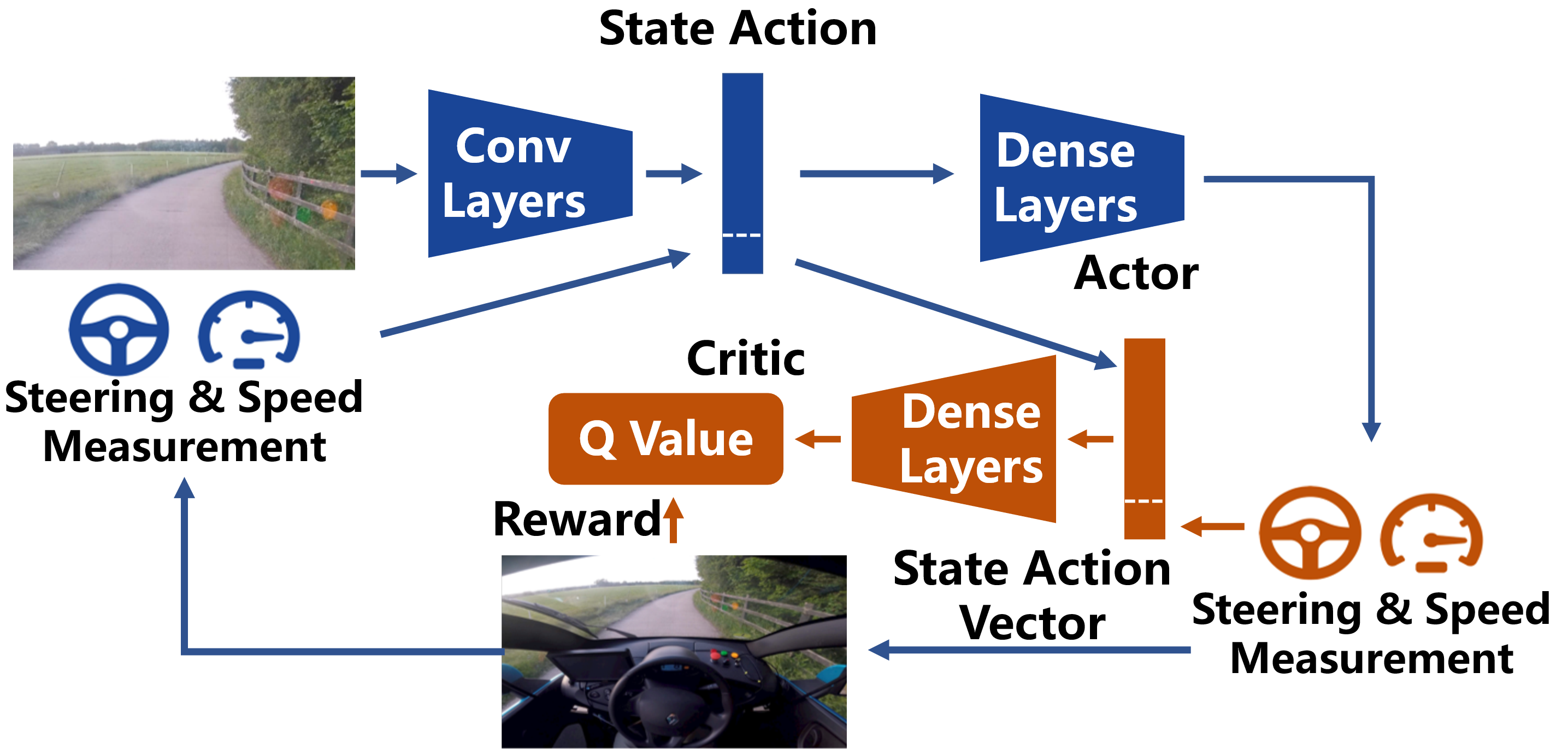}
    \caption{The actor-critic algorithm used to learn a policy and value function for driving proposed in \cite{RL-kendall2019}.}
    \label{fig:RL-kendall2019}
\end{figure}

\subsubsection{Policy-based Reinforcement Learning}

The value-based approach is limited to providing discrete commands. However, autonomous driving is a continuous process, continuous commands within an uninterrupted span can be controlled at a fine-grained level. Therefore, the continuous approach is better for vehicle control. Policy-based methods hold the potential for high ceilings in high-dimensional action spaces with continuous control commands. These methods exhibit superior convergence and exploration than value-based methods.

The execution of RL on real-world IVs is a challenging assignment. Kendall et al. \cite{RL-kendall2019} implement the Deep Deterministic Policy Gradient (DDPG) \cite{RL-ddpg-lillicrap2015} algorithm on an actual intelligent vehicle, performing all exploration and optimization on-board, as shown in Fig. \ref{fig:RL-kendall2019}. Monocular images are the only input, the agent learns the lane-following policy and achieves human-level performance in a 250m road test. This work marks the first application of implementing deep reinforcement learning on a full-sized autonomous vehicle. To further enhance driving safety and comfort, Wang et al. \cite{RLTsy6} introduce an innovative method for IVs based on the lane-change policy of human experts. This method can be executed on single or multiple vehicles, facilitating smooth lane changes without the need for V2X communication support.

To alleviate the challenge of autonomous driving on congested roads, Saxena et al. \cite{RL-saxena2020} employ the proximal policy optimization (PPO) algorithm \cite{rl-ppo-schulman2017} to learn a control policy in a continuous motion planning space. Their model implicitly simulates interaction with other vehicles to avoid collisions and enhance passenger comfort.  Building on this work, Ye et al. \cite{ye2020} leverage PPO to learn an automated lane change policy on real highway scenarios. Taking the ego vehicle and its surrounding vehicle states as input, the agent learns to avoid collisions and to drive in a smooth manner. \textcolor{black}{Several other studies \cite{guan2020, wuyuanqing2021} have also demonstrated the efficacy of PPO-based RL algorithms in end-to-end autonomous driving, since PPO can provide better performance for both the efficiency of policy learning and the diversity during trajectory exploring.}

Training a policy from scratch in RL is frequently time-consuming and difficult. Combining RL with other methods such as imitation learning (IL) and curriculum learning may serve as a viable solution. Liang et al. \cite{RL-liang2018cirl} combine IL and DDPG together to alleviate the problem of low efficiency in exploring the continuous space, an adjustable gating mechanism is introduced to selectively activate four different control signals, which allows the model to be controlled by a central one.  Tian et al. \cite{RLTsy2} leverage an RL method of learning from expert experience to implement trajectory-tracking tasks, which are trained in two steps, an IL method adopted in \cite{IL7} and a continuous, deterministic, model-free RL algorithm to further fine-turn the method. 

\begin{figure}
    \centering
    \includegraphics[width=0.8\linewidth]{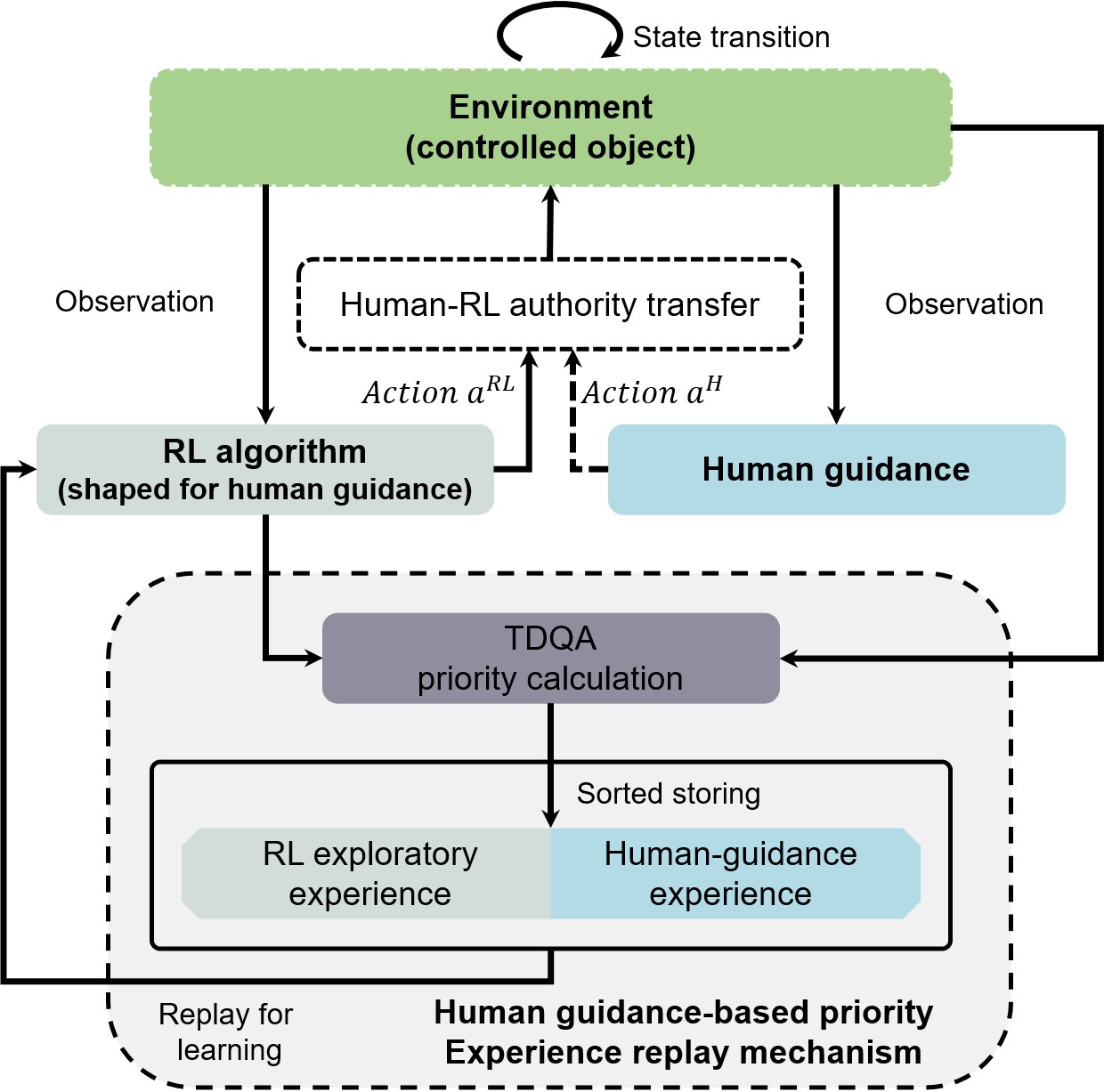}
    \caption{Framework of the proposed human guidance-based RL algorithm \cite{RLTsy4}.}
    \label{fig:LvChen}
\end{figure}

To address the learning efficiency limitations of RL methods, Huang et al. \cite{RLTsy3} devise a novel method, which incorporates human prior knowledge in RL methods. When confronted with the long-tail problem of autonomous driving, many researchers have turned their perspective to the exploitation of expert human experience. Wu et al. \cite{RLTsy4} propose a human guidance-based RL method which leverages a novel prioritized experience replay mechanism to improve the efficiency and performance of the RL algorithm in extreme scenarios, the framework of the proposed method is shown in Fig \ref{fig:LvChen}. This method is validated in two challenging autonomous driving tasks and achieves a competitive result. \textcolor{black}{Therefore, improving the performance of driving tasks may require the combination of multiple methods and the design of task-specific training methods.}

\subsubsection{Hierarchical Reinforcement Learning}

RL methods have shown great promise in various domains, however, these methods are often criticized for difficult training. Especially in the autonomous driving field, non-stationary scenarios and high-dimensional input data cause intolerable training hours and resource usage \cite{JAS3}. Hierarchical reinforcement learning (HRL) decomposes the total problem into a hierarchy of subtasks, and each subtask has its own goal and policy. The subtasks are organized in a hierarchical manner, with higher-level subtasks providing context and guidance for lower-level ones. This hierarchical organization allows the agent to focus on smaller subproblems, reducing the complexity of the learning problem and making it more tractable. 

Forcing the lane-changing task, Chen et al. \cite{chen_hrl_2019} propose a two-level method. The high-level network learns policies for deciding whether to execute a lane change action, while the low-level network learns policies for executing the chosen commands. \cite{HRL-shi2019} and \cite{li_hrl_2021} also present a two-stage HRL methodology based on \cite{chen_hrl_2019}, where \cite{HRL-shi2019} needs to employ the pure pursuit to track the output trajectory points, and \cite{li_hrl_2021} integrates position, velocity and heading of ego-vehicle to further improve the performance of the low-level controller.  All these proposed methods provide a promising solution for developing robust and safe autonomous driving systems.

\begin{figure}[b]
    \centering
    \includegraphics[width=0.98\linewidth]{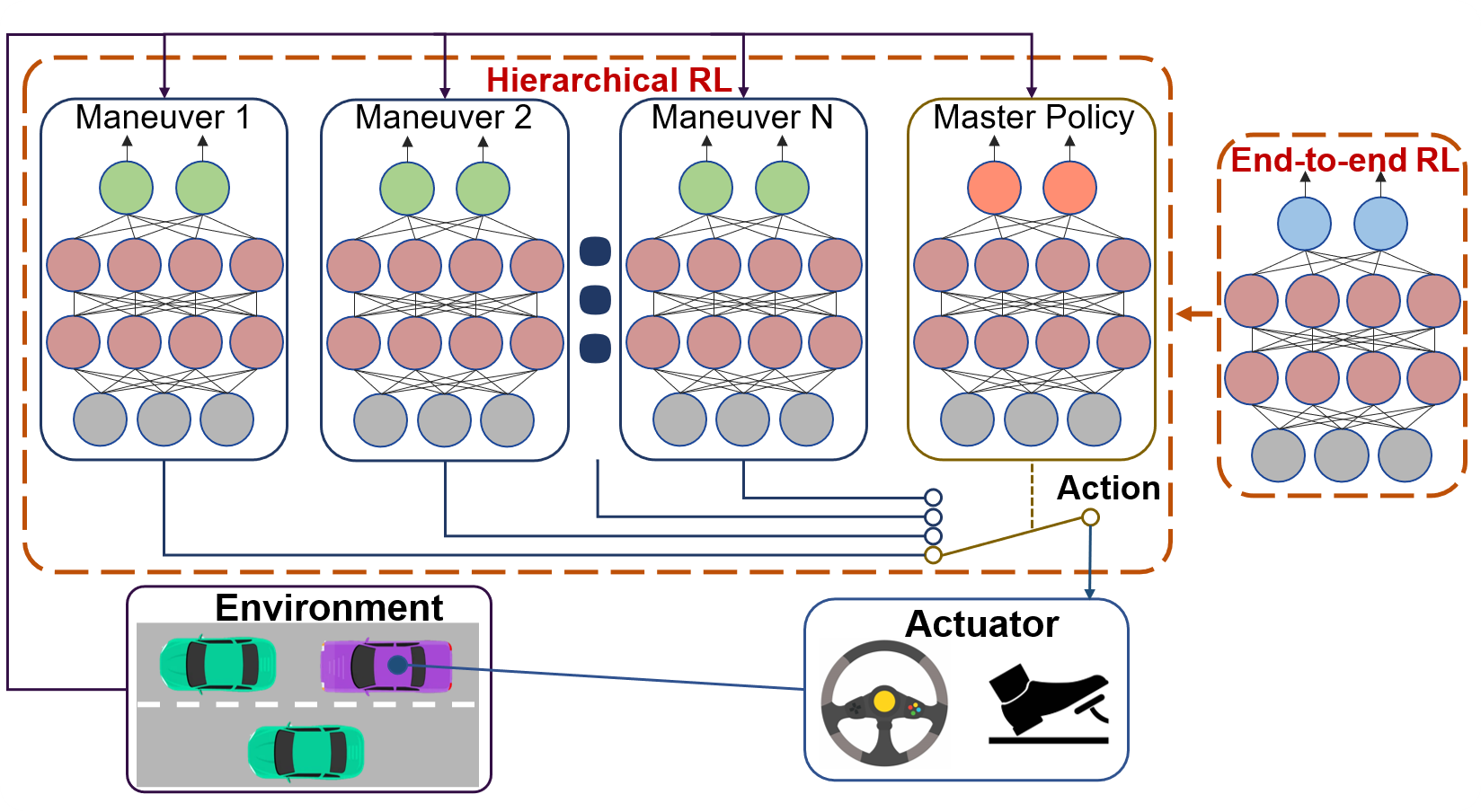}
    \caption{The framework of hierarchical reinforcement learning (HRL) for self-driving decision-making proposed in \cite{duan2020hierarchical}.}
    \label{fig:HRL-Duan2020}
\end{figure}

The generalizability of HRL is a hot research point. Lu et al. \cite{HRL-Lu2020} propose an HRL approach for autonomous decision-making and motion planning in complex dynamic traffic scenarios, as shown in Fig. \ref{fig:HRL-Duan2020}. The approach consists of a high-level layer and a low-level planning layer, the high-level layer leverages a kernel-based least-squares policy iteration algorithm with uneven sampling and pooling strategy (USP-KLSPI) to solve the decision-making problems. Duan et al. \cite{duan2020hierarchical} divide the whole navigation task into three models. The master policy network is trained to select the appropriate driving task, this policy greatly enhances the generalizability and effectiveness of the model. For the purpose of further improving decision quality in complex scenarios, Cola-HRL \cite{HRLTsy2_2022} is presented based on \cite{duan2020hierarchical}, this method consists of three main components: a high-level planner, a low-level controller, and a continuous-lattice representation of the state space. \textcolor{black}{The results show that the Cola-HRL outperforms other SOTA methods for making high-quality decisions in various scenarios.}


\subsubsection{Multi-Agent Reinforcement Learning}
In real scenarios,  diverse traffic participants are commonly present, and their interactions can have a significant impact on the policy of each other \cite{xu2022bridging}. In the single-agent system, the behavior of other participants is usually controlled based on pre-defined rules, and the predicted behavior of the agent may overfit the other participants, thus leading to a more deterministic policy other than in a multi-agent one \cite{JAS2}. Multi-Agent Reinforcement Learning (MARL) is designed to learn the decision-making policies of multiple agents in the environment. \textcolor{black}{Decentralized partially observable Markov decision processes (DEC-POMDPs) are a typical formalization of MARL, as in many real-world domains, it is not possible for agents to observe all features of the environment state, and all agents interact with the environment in a decentralized way.} Furthermore, the state space expands exponentially with the number of agents, making it more challenging and slow to train a multi-agent system (MAS) \cite{xu2022model}.

To reduce the impact of ``the dimension explosion", some effective learning schemes are proposed. Kaushik et al. \cite{Kaushik19} use a simple parameter-sharing DDPG to train the agent for two distinct tasks. By injecting the task into the observation space as a command, the same agent can act both competitively or cooperatively. Wang et al. \cite{wang2020multi} train autonomous agents in three scenarios: a ring network, a figure-of-eight network, and a mini city with various scenarios. Graph information sharing between each agent is integrated in the approach with PPO for continuous action generation, and vehicle communication is permitted within a certain range.

Although RL has been widely studied for lane-changing decision makings, those studies are mainly focused on a single-agent system. MARL methods provide a global perspective on multi-vehicle control. Zhou et al. \cite{zhou2022multi} formulate the lane-changing decision-making of multiple autonomous vehicles coexisting with human-driven vehicles in a mixed-traffic highway scenario. Beyond simple tasks, MARL approaches have great potential to solve decision and planning problems in complex scenarios. Chen et al. \cite{chen2021deep} train agents to evade collisions in a time-critical merging highway scenario. The agents observe the locations and the velocities of the surrounding vehicles and then select corresponding actions.

\begin{figure}
    \centering
    \includegraphics[width=0.88\linewidth]{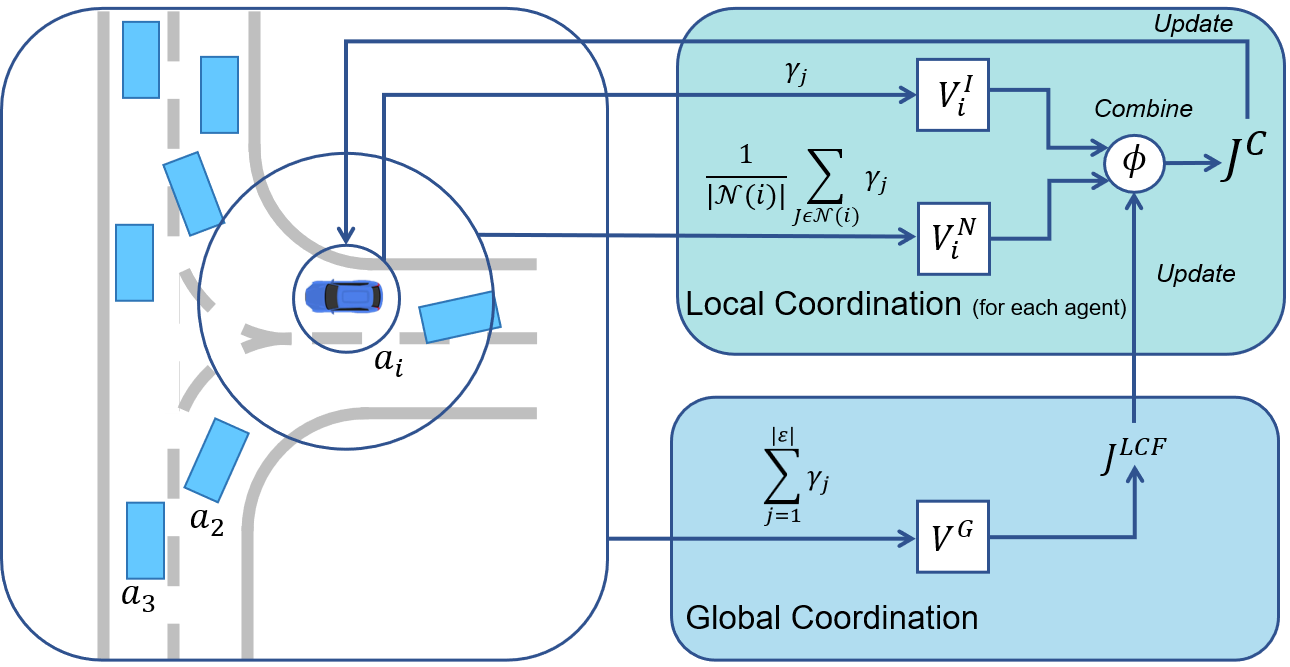}
    \caption{The framework of the CoPO method proposed in \cite{peng2021learning}: the Local Coordination Factor (LCF) describes an agent’s preference of being selfish, cooperative, or competitive. A Local Coordination for each policy and a Global Coordination to update global LCF are both performed during training.}
    \label{fig:CoPO}
\end{figure}

Credit assignment is vital for policy learning in cooperative multi-agent scenarios. Han et al. \cite{han2022} introduce an effective reward reallocating mechanism to motivate stable cooperation among IVs using a cooperative policy learning algorithm with Shapley value reward reallocation. \textcolor{black}{Each vehicle’s states include position, velocity, acceleration, image and point clouds captured by the onboard camera and LiDAR sensors respectively.} The experimental results demonstrate significant improvement of the mean episode system reward in connected autonomous vehicles. \textcolor{black}{Instead of reallocating rewards between agents, Peng et al. \cite{peng2021learning} incorporate the social psychology principle to learn the neural controller of Self-Driven Particles (SDP) system, in which each constituent agent is self-interested and the relationship between them is constantly changing. The proposed method, Coordinated Policy Optimization (CoPO), performs local  coordination between the agent and its neighbor vehicles global coordination, as shown in Fig \ref{fig:CoPO}. Taking raw LiDAR data as input and continuous actuator signals as output, experiments demonstrate that the proposed method outperforms other MARL methods across three main metrics: success rate, safety, and efficiency. This work adopts 5 typical simulated traffic scenarios, which are still far from emulating the complexity of real-world traffic scenes and lack other traffic participants, e.g., pedestrians and traffic lights.} \textcolor{black}{Compared with the monocular camera, LiDAR can provide sufficient range information, making agents interact within a safer distance. so these works \cite{Kaushik19,han2022,peng2021learning} achieve more generalization results.}

\textcolor{black}{Although RL is an appealing way to make the agent learn by trial-and-error in the environment without expert instructions, most RL methods suffer from poor sample efficiency. With the use of neural networks for deep representation learning and function approximation in the domain of RL, interpretability still remains a challenge.}

\begin{table*}
\newcommand{\tabincell}[2]{\begin{tabular}{@{}#1@{}}#2\end{tabular}}  
\centering
\caption{MAIN APPROACHES FOR MOTION PLANNING IN AUTONOMOUS DRIVING BASED ON DEEP REINFORCEMENT LEARNING.}
\label{tab:reinforcementlearning}
\begin{tabular}{m{2.25cm} m{1.8cm}<{\centering} m{2.5cm}<{\centering}  m{4cm}<{\centering}  m{2.4cm}<{\centering}  m{3cm}<{\centering}}
\hline
\rowcolor[rgb]{0.937,0.937,0.937} \multicolumn{1}{c}{\textbf{Article}} & \textbf{Method} & \multicolumn{1}{c}{\textbf{Observation}}      & \multicolumn{1}{c}{\textbf{Output}} & \multicolumn{1}{c}{\textbf{Scenario}}         & \multicolumn{1}{c}{\textbf{\textbf{Simulator}}}        \\ 


\hline

Wolf et al. \cite{RL-wolf2017}    & \textbf{Value-based}, DQN  & front cam    & discrete Steering angle    & lane keeping       & Gazebo \\

Alizadeh  et al. \cite{Alizadeh2019}  & \textbf{Value-based}, DQN  & \tabincell{c}{ relative distance \& \\ velocity value} & trajectory points  & lane change     & \tabincell{c}{Self-made environment}\\

Ronecker  et al. \cite{Ronecker2019} & \textbf{Value-based}, DQN  & \tabincell{c}{ relative distance \& \\ velocity value}   & trajectory points  & \tabincell{c}{lane change,\\highway strategy}   & \tabincell{c}{Self-made environment}\\

Li et al. \cite{RLTsy7} & \textbf{Value-based}, DQN  & front cam  & discrete lane change action   & \tabincell{c}{lane change \& \\city strategy}    & CARLA \\

Mo et al. \cite{RLTsy9} & \textbf{Value-based}, DQN  & front cam  & \tabincell{c}{discrete acceleration \& \\ lane change action}   & \tabincell{c}{overtakeing \& \\highway strategy}     & SUMO \\

Kendall  et al. \cite{RL-kendall2019}   & \textbf{Policy-based}, DDPG    & front cam    & \tabincell{c}{continuous steering angle \& \\ speed setpoint}    & lane keeping        & Unreal Engine 4    \\

Wang et al. \cite{RLTsy6}   & \textbf{Policy-based}, DDPG, DQN    & front cam    & discrete lane change action   & lane change      & Self-made environment   \\

Saxen  et al. \cite{RL-saxena2020}   & \textbf{Policy-based}, PPO   & lane based grid   & \tabincell{c}{continuous acceleration \& \\steering angle}   & highway kinematic	    & Open source simulator \\

Ye et al. \cite{ye2020}  & \textbf{Policy-based}, PPO   & \tabincell{c}{relative distance \& \\ velocity}   & discrete lane change action   & lane change	    & SUMO \\

Liang  et al. \cite{RL-liang2018cirl}	& \textbf{Policy-based}, DDPG  	& front cam, Speed	  & \tabincell{c}{continuous steering angle, \\ acceleration, braking}	 & \tabincell{c}{navigation}	      & CARLA \\

Tian  et al. \cite{RLTsy2} 	  & \textbf{Policy-based}, BC, DDPG    & vehicle kinematic	  & \tabincell{c}{continuous steering angle \& \\vehicle speed}  & path tracking      & Carsim/Simulink \\

Huang  et al. \cite{RLTsy3}   & \textbf{Policy-based}, BC, AC     & BEV images    &    \tabincell{c}{continuous target speed \& \\ discrete lane change action}	  & \tabincell{c}{unprotected left turn, \\roundabout}      & SMARTS \\

Wu  et al. \cite{RLTsy4}   & \textbf{Policy-based}, PHIL-TD3	& BEV semantic graph	& \tabincell{c}{ continuous steering angle \& \\accelerating}  & left-turn, congestion	   	  & CARLA \\

Chen et al. \cite{chen_hrl_2019}  & \tabincell{c}{\textbf{HRL}, \\AC, DQN}     & front cam  & trajectory points  & lane change    & TORCS \\

Shi	 et al. \cite{HRL-shi2019}  & \tabincell{c}{\textbf{HRL}, \\DQN}      & relative distance \& velocity  & \tabincell{c}{discrete lane change action \& \\ continuous acceleration}  & lane change   & Self-made environment\\

Li et al. \cite{li_hrl_2021}  & \tabincell{c}{\textbf{HRL},\\ DQN}  & scenario state  & \tabincell{c}{discrete speed \& \\ steering angle}    & INTERACTION dataset & OpenAI GYM toolkit\\

Duan  et al. \cite{duan2020hierarchical}  &\textbf{HRL}  & policy-specific dynamics  & \tabincell{c}{discrete speed \& \\ steering increment}	 &\tabincell{c}{lane change}	 	& Self-made environment  \\

Lu  et al. \cite{HRL-Lu2020}  & \textbf{HRL}, USP-KLSPI  & 14-DOF dynamics	 & \tabincell{c}{discrete speed \& \\steering action}   & \tabincell{c}{lane merging}	  & Matlab \\

Gao et al. \cite{HRLTsy2_2022}  &\tabincell{c}{\textbf{HRL}, \\DDPG, CNN}  & BEV perception data, HD-Map  & \tabincell{c}{ continuous speed \& \\ steering angle} 	 &navigation 	  	& Real-world HD-maps  \\

Kaushik et al. \cite{Kaushik19}  & \textbf{MARL}, DDPG  & \tabincell{c}{vehicle kinematics,\\LiDAR}  & \tabincell{c}{continuous speed \& \\ steering angle}   & highway navigation    &TORCS  \\

Wang et al. \cite{wang2020multi}  &\tabincell{c}{ \textbf{MARL}, \\PPO}    & relative distance \& velocity   & continuous acceleration & road networks   &Flow  \\

Zhou et al. \cite{zhou2022multi}    & \tabincell{c}{\textbf{MARL},  \\MA2C}    & relative distance \& velocity   &  discrete acceleration & lane change action    & Highway-env  \\

Chen et al. \cite{chen2021deep}  & \tabincell{c}{\textbf{MARL}, \\MA2C}    & relative distance \& velocity   & \tabincell{c}{discrete acceleration \& \\ lane change action}  & lane merging     & Highway-env  \\

Han et al. \cite{han2022}    & \tabincell{c}{\textbf{MARL},\\Reward\\Reallocation}    & \tabincell{c}{front cam, LiDAR,\\vehicle kinematic}   & discrete lane change action  & mixed traffic  & CARLA  \\

Peng et al \cite{peng2021learning}      & \tabincell{c}{\textbf{MARL},  \\CoPO}   & \tabincell{c}{continuous ego state \& \\LiDAR}   & \tabincell{c}{continuous acceleration \& \\ steering angle values} & multi scenarios   & MetaDrive  \\

\hline
\end{tabular}
\end{table*}

\subsection{Parallel Learning}


Planning methods in autonomous driving are constrained by several challenges. Pipeline planning methods couple numerous human-customized heuristics, which leads to inefficient computation and low generalization. Imitation learning (IL) methods require considerable volume and diverse distribution of expert trajectories, while reinforcement learning (RL) methods demand significant computational resources. Consequently, the presence of these limitations impedes the widespread implementation of autonomous driving.

\begin{figure}[b]
    \centering
    \includegraphics[width=0.98\linewidth]{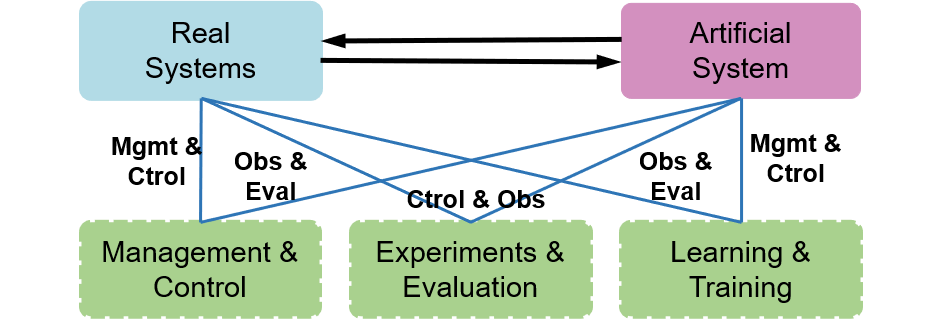}
    \caption{The framework of parallel system theory proposed in \cite{ParallelACP}.}
    \label{fig:ACP}
\end{figure}

\begin{table*}[b]
\centering 
\label{tab:parallelmethods}
\caption{The survey about the parallel system theory and its sources and derived algorithms.}
\begin{tabular}{c c m{13cm}}
\hline
\rowcolor[HTML]{EFEFEF} 
Method                 & Year  & \multicolumn{1}{c}{\cellcolor[HTML]{EFEFEF}Detail}                                                                                                                                                                       \\ \hline
CPS                    & 1990s & Proposing a multi-dimensional intelligent technology framework, based on big data, internet of things, and large computing, the organic integration and deep collaboration of computing, communication and control (3C). \\
CPSS \cite{Cpss}                  & 2000  & Integrating social signals and relationships into CPS, leveraging the human, data and information of the social network to break through the various limitations of the real world.                                      \\
Parallel System Theory \cite{ParallelACP} & 2004  & Integrating artificial societies (A), computational experiments (C) and parallel execution (P), and provide effective tools for control and management of complex systems.                                               \\
Parallel Learning \cite{ParallelLearning}      & 2017  & Proposing a new framework of machine learning theory, parallel learning, which incorporates and inherits many elements from various existing machine learning theories.                                                  \\
Parallel Vision \cite{parallelvsion}        & 2017  & Introducing the parallel system theory into the computer vision area and constructing a novel research method for perception and understanding of complex driving scenarios.                                             \\
Parallel Driving \cite{ParallelDriving}       & 2019  & Constructing an advanced and unified framework for autonomous driving that includes operation management, online condition management and emergency disengagement.                                                       \\
Parallel Planning \cite{ParallelPlanning2}      & 2019  & Constructing a deep planning method that integrates a convolutional neural network and a Long short-term memory module to improve the generalization and robustness of planning models in intelligent vehicles.           \\
Parallel Testing \cite{paralleltest}       & 2019  & Proposing a closed-loop testing framework, which implements more challenging scenarios to accelerate evaluation and development of autonomous vehicles.                                                                  \\ \hline
\end{tabular}
\end{table*}


In response to the various problems in planning methods, virtual-real interaction provides a proven solution \cite{JAS_FeiYue}. Cyber-physical-systems (CPS) based intelligent control can facilitate interactions and integration between physical and cyberspaces but are not considering human and social factors in systems. In reply, many researchers have added social factors and artificial information to the CPS to form the cyber-physical-social systems (CPSS). In CPSS, the 'C' stands for two dimensions: the information system in the real world and the virtual artificial system defined by software. The 'P' refers to the traditional real system. The 'S' includes not only the human social system but also the artificial system based on the real world.

CPSS enables virtual and real systems to interact, feedback, and promote each other. The real system provides valuable datasets for the construction and calibration of the artificial system, while the artificial system directs and supports the operation of the real system, thus achieving self-evolution. Due to the advantages of virtual-real interaction, CPSS provides a new verification method for end-to-end autonomous driving.

Based on CPSS, Fei-Yue Wang \cite{ParallelACP} proposes the concept of parallel system theory in 2004, as shown in Fig. \ref{fig:ACP}, the core concept of which is the ACP method, an organic combination of artificial societies (A), computational experiments (C) and parallel execution (P). Over the past two decades, the research system of parallel system theory has been enriched and improved by a large number of implementations in practice \cite{ITSM}, such as parallel intelligence \cite{parallelintelligence}, parallel control \cite{parallelcontrol, Lu_ParallelChontrol}, parallel management \cite{ParallelFactory}, parallel transportation \cite{paralleltransportation}, parallel driving \cite{ParallelDriving, Paralleljas}, parallel tracking \cite{Lu_parallelTracking}, parallel testing \cite{paralleltest}, parallel vision \cite{parallelvsion} and so on. The survey about the methods proposed in this section is shown in Table \ref{tab:parallelmethods}.





\begin{figure}
    \centering
    \includegraphics[width=0.88\linewidth]{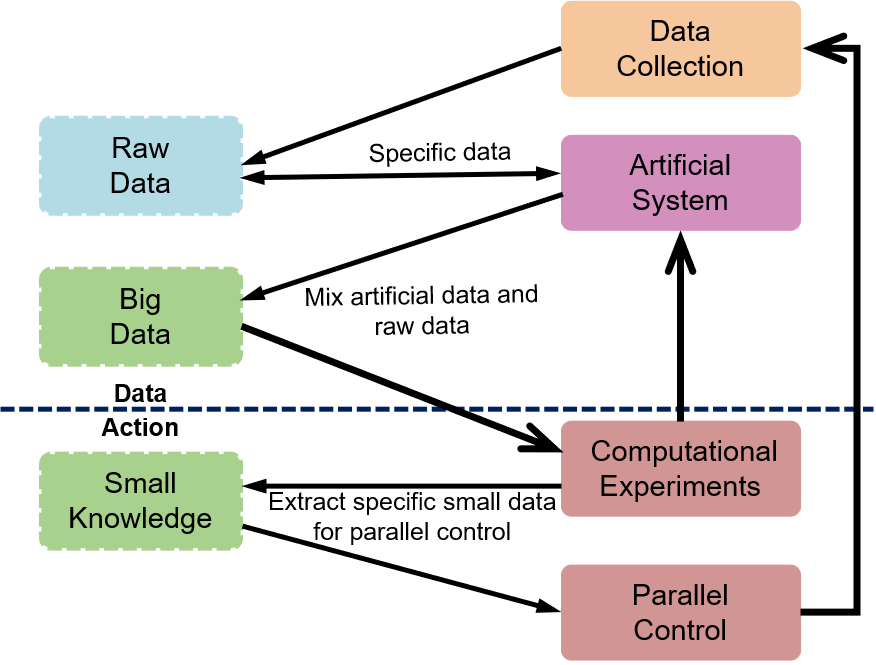}
    \caption{ The theoretical framework of parallel learning proposed in \cite{ParallelLearning}. (The part above the dashed line focuses on big data preprocessing using artificial systems; the part beneath the dashed line focuses on computational experiments. The thin arrows represent either data generation or data learning; the thick arrows present interactions between data and actions.)}
    \label{fig:ParallelLearning}
\end{figure}

In order to further expand the learning capabilities of neural networks, and to address the challenges of IL and RL, Li et al. \cite{ParallelLearning} propose a basic framework for parallel learning based on the parallel system theory as shown in Fig. \ref{fig:ParallelLearning}. In the action phase, parallel learning \cite{ParallelLearning} follows the RL paradigm, employing state transfer to represent the movement of the model, learning from big data, and storing the learned policy in the state-transition function. Notably, parallel learning capitalizes on computational experimentation to refine the policy. Through feature extraction methods, small knowledge can be applied to specific scenarios or tasks, and used for parallel control. Here, ``small" refers to specific and intelligent knowledge for the particular problem, rather than denoting the magnitude of knowledge.

\begin{figure}[b]
    \centering
    \includegraphics[width=0.88\linewidth]{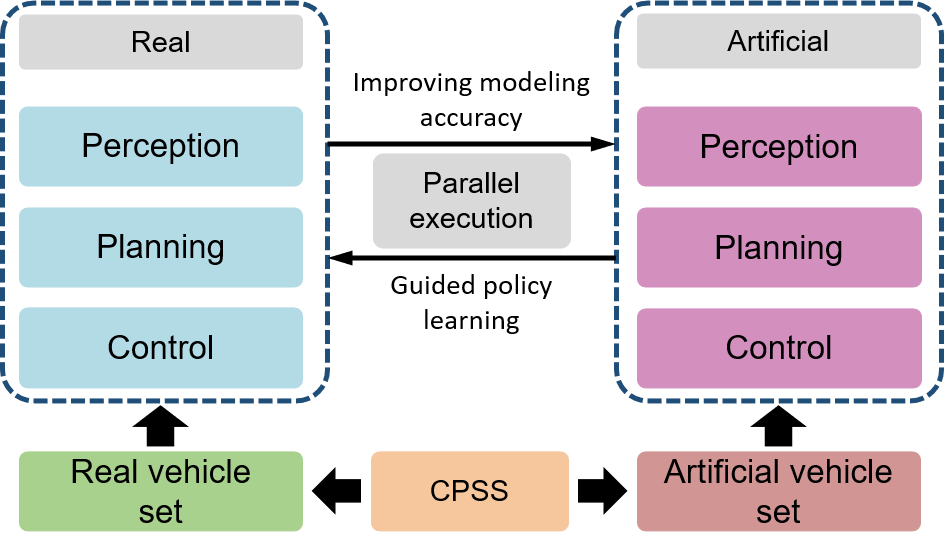}
    \caption{The theoretical framework of the parallel driving proposed in \cite{ParallelDriving2}.}
    \label{fig:ParallelPlanning}
\end{figure}

An innovative training approach based on parallel learning \cite{ParallelLearning} presents an alternative solution for problem-solving in fully end-to-end autonomous stacks. As shown in Fig. \ref{fig:ParallelPlanning}, Wang et al. \cite{ParallelDriving2}  introduce a parallel driving framework, a unified approach for ITS and IVs. The framework directly bridges expert trajectories and control commands to compute the most optimal policy for specific scenarios. Plenty of expert trajectories are collected from real scenarios, and a neural network is employed to learn all these trajectories, inputs and outputs of this network are destination state and control signals. From the viewpoint of parallel learning, this is a self-labeling process, and the process significantly alleviates the data hunger of end-to-end methods.


In order to handle the integrated data from the artificial system and computational experiment, a new theory is proposed, named parallel reinforcement learning (PRL), which combines the parallel learning and deep reinforcement learning approaches. Liu et al. \cite{ParallelDriving} integrate digital quadruplets with parallel driving. This framework defines the physical vehicle, the descriptive vehicle, the predictive vehicle, and the prescriptive vehicle. Based on the description of digital quadruplets, three virtual vehicles can be defined as three ``guardian angels" for the physical vehicle, playing different roles to make the IVs safer and more reliable in complex scenarios.

\begin{figure}[t]
    \centering
    \includegraphics[width=0.8\linewidth]{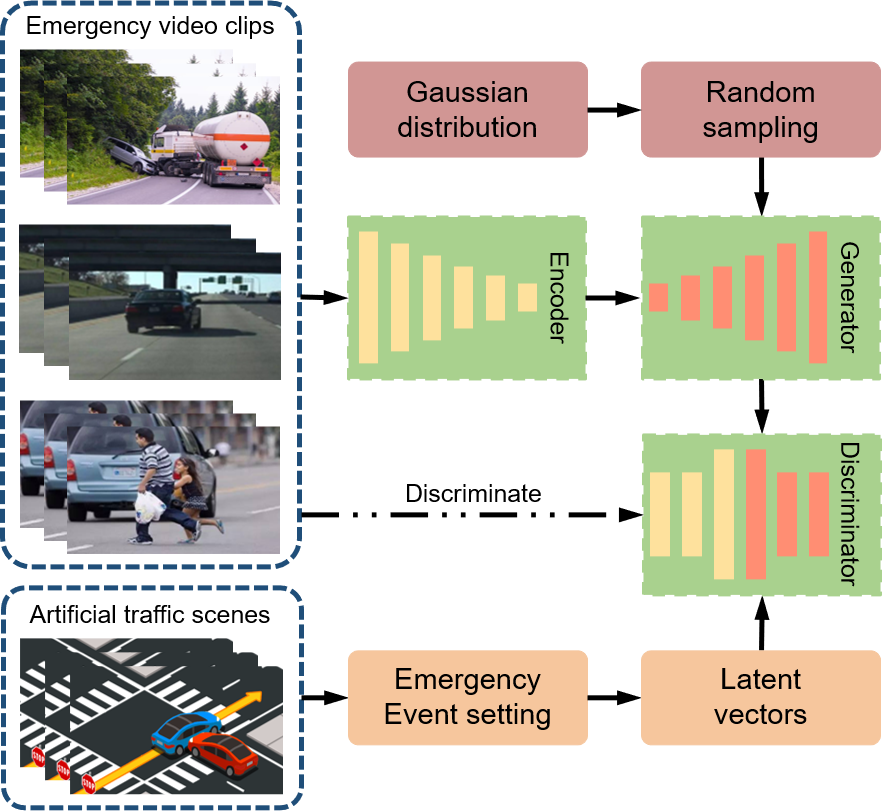}
    \caption{Hybrid model of combining the variational auto-encoder (VAE) and the generative adversarial network (GAN) for predicting and generating potential emergency image sequences proposed in \cite{ParallelPlanning2}.}
    \label{fig:ParallelDrivining}
\end{figure}

Planning is one of the most significant components of autonomous driving. As a concrete implementation of parallel driving, Chen et al. \cite{ParallelDriving, ParallelDriving2} propose a parallel planning framework for end-to-end planning, which constructs two customized approaches to solve emergency planning problems in specific scenarios. For the data-insufficient problem, parallel planning leverages artificial traffic scenarios to generate expert trajectories based on the pretrained knowledge from reality, as shown in Fig. \ref{fig:ParallelDrivining}. For the non-robustness problem, parallel planning utilizes a variational auto-encoder (VAE) and a generative adversarial network (GAN) to learn from virtual emergencies generated in artificial traffic scenes. For the learning inefficient problem, parallel planning learning policy from both virtual and real scenarios, and the final decision is determined by analysis of real observations. Parallel planning is able to make rational decisions without a heavy calculation burden when an emergency occurs.

The parallel system theory provides an effective tool for the control and management of complex systems, especially in the autonomous control field, parallel driving effectively alleviates the shortage of data, inefficient learning, and poor robustness for end-to-end planning models.

\begin{table*}
\caption{Datasets and related descriptions for the autonomous driving dataset.}
\label{tab:dataset}
\centering
\begin{tabular}{cccc} 
\hline
\rowcolor[rgb]{0.937,0.937,0.937} Dataset & Year & Sensors                                       & Scenarios                        \\ 
\hline
KITTI \cite{kitti}                                     & 2013 & 4 cameras; 1 LiDAR                            & City; Countryside; Highway       \\
Comma.ai \cite{commonai}                                  & 2016 & 1 monocular camera; 1 point grey
  camera     & Highway scenarios                \\
Oxford
  RobotCar \cite{oxfod}                         & 2016 & 6 Cameras; 3 LiDARS; Speed; GPS; INS                 & City; Contain weather changed    \\
Mapillary
  Vistas \cite{mapillary}                        & 2017 & Image devices                                 & Street Scenarios                  \\
nuScenes \cite{nuscenes}                                  & 2019 & 6 Cameras; 5 Radars; 1 LiDAR                  & Street Scenarios                  \\
ApolloScape \cite{apolloscape}                               & 2019 & 2 Cameras; 2 LiDAR; GPS; IMU                  & Street Scenarios                  \\
Waymo
  Open Dataset \cite{waymo}                      & 2019 & 5 Cameras; 5LiDAR;~                           & 1150 Street Scenarios            \\
BDD100K \cite{BDD}                                   & 2020 & 1 Camera; GPS; IMU                            & Street scenarios in 4 cities     \\
A2D2 \cite{A2D2}                                      & 2020 & 6 Cameras; 5 LiDAR; GPS; IMU                    & 360° Street Scenarios             \\
Automine \cite{Automine}                                  & 2021 & 2 Cameras; 1 LiDAR; GPS; IMU                    & The first open-pit mine dataset  \\
AI4MARS \cite{Mars}                                  & 2021 & 2 Cameras                                  & The first large-scale dataset in Mars  \\

SODA10M \cite{huawei}                                  & 2021 & 1 Camera                                  & City Scenarios in 31 cities with all kinds of weathers   \\

SUPS \cite{SUPS}                                      & 2022 & 6 Cameras; 1 LiDAR; GPS; IMU                   & Underground parking scenarios    \\
DRIVERTRUTH \cite{drivetruth}                               & 2022 & 1 Camera; 1 LiDAR; GPS; IMU;
  Control signal & City Scenarios based-on CARLA    \\

ROAD \cite{PAMI}                                      & 2023 & 1 Camera             & Scenarios in \cite{oxfod} for road event detection    \\
\hline
\end{tabular}
\end{table*}

\section{EXPERIMENT PLATFORM \label{experiments}}
Testing IVs in real systems often comes with potentially fatal safety risks. Therefore, algorithms in autonomous driving are often evaluated in artificial systems with the utilization of open-source datasets and simulation platforms \cite{OpenCDA}.

\subsection{Dataset}
The end-to-end method leverages widely available large-scale datasets of human driving to be trained to approximate human standards. Consequently, the training process requires a large amount of data from driving scenarios. The magnitude, abundance, and distribution of the dataset directly affect the safety, robustness, and generalization of the trained model. Though constructing and assembling novel datasets for IVs is time-consuming, numerous generic and influential datasets are available for research, such as Comma.ai \cite{commonai}, Bdd100K \cite{BDD}, A2D2 \cite{A2D2}, Automine \cite{Automine}, DriverTruth \cite{drivetruth} and Sups \cite{SUPS}, most of the famous dataset is shown in Table. \ref{tab:dataset}.

KITTI \cite{kitti} is a pioneer in this field and also the most famous autonomous driving dataset. Thanks to its good task scaling, KITTI now covers a wide range of basking perception tasks, such as object detection, sceneflow, depth estimation, tracking and so on.

Comma.ai \cite{commonai} enriches the diversity of data by additionally collecting localization information and control signals, so it can be implemented for more tasks, for example, localization and planning.

BDD100K \cite{BDD} and SODA10M \cite{huawei} alleviate diversity and volume problems by constructing large-scale simulation scenarios, both of them collect several urban scenarios under various weather conditions in more than 31 cities, they also come with a rich set of labels: scene tagging, object bounding box, lane marking, drivable area, full-frame semantic and instance segmentation, multiple object tracking, and multiple objects tracking.

A2D2 \cite{A2D2} is a commercial-grade dataset that is well-suited for diverse perception tasks, bridging the gap between public datasets which are deficient in comprehensive vehicular information. Compared with previous datasets, it provides a 360° point cloud perception field by 5 LiDARs to enable full scene perception for autonomous driving.

The following dataset provides traffic scenarios that are distinct from previous ones. Automine \cite{Automine} constructs the pioneering open-pit mine dataset for IVs, comprising 18 hours of transportation videos and localization information gathered from 6 open-pit mines.  The distinctive features of open-pit mines, such as uneven and rough terrain, intense light, and copious dust, pose significant challenges. The Automine serves as a valuable resource to address the gaps in the open-pit mine dataset, and supports the advancement of autonomous mining technology. AI4MARS \cite{Mars} proposes another interesting large-scale dataset, which consists of 35,000 semantic segmentation full images of the surface of Mars.

\textcolor{black}{Currently, datasets play a crucial role in training and validating IV methods, supporting the fundamental groundwork necessary for implementing autonomous driving.}

\subsection{Simulation Platform}
Testing autonomous driving algorithms in real-world scenarios is often accompanied by significant potential risks, simulation testing shows a smart method to validate algorithms that can speed up testing due to its low cost and high safety.


Many autonomous driving simulation platforms have been developed with open-source code and protocols, which are available for the testing of algorithms in autonomous driving. SUMO \cite{SUMO}, an open-source and microscopic traffic simulation platform, developed by the German Aerospace Center, offers a powerful validation platform for large-scale transportation algorithms. It is equipped with a well-designed interface that supports a broad range of data formats. Owing to its superior features, SUMO has been one of the earliest and most extensively utilized simulators. Moreover, Apollo \cite{apolloscape} and Autoware \cite{autoware} not only provide a simulation platform for validating algorithms but also equip open-source algorithms for each task, providing developers with a complete development-validation-deployment chain.

In the context of the ego-vehicle autonomous driving method, CARLA \cite{carla} offers a suitable answer. It is an open-source simulator for urban autonomous driving scenarios, which facilitates the development, training, and validation of the underlying urban autonomous driving system.

In the field of the multi-vehicle interaction method, TORCS \cite{torcs}  provides an open racing car simulator with over 50 diverse vehicle models and more than 20 racing tracks. Furthermore, it has the ability to race against 50 vehicles simultaneously, making it a valuable tool for research in this field. MetaDrive \cite{metadrive} proposes an open-source platform to support the research of generalizable reinforcement learning algorithms for machine autonomy. It is highly compositional and capable of generating an infinite number of diverse driving scenarios through both procedural generation and real data importing. The other simulation platforms and their related descriptions are shown in Table. \ref{tab:simulator}.



\begin{table*}
\caption{Simulation platforms and related descriptions for autonomous driving based on visual perception.}
\label{tab:simulator}
\centering
\begin{tabular}{cc m{12cm}} 
\hline
\rowcolor[rgb]{0.937,0.937,0.937} \textbf{Platform} & \textbf{Latest Version} & \multicolumn{1}{c}{\textbf{Description}}                                                                                                                        \\ 
\hline
PTV 
  Vissim                                        & V2023                   & Traffic simulation platform
  focused on complex intersection design and active traffic management.                                                             \\
VTD                                                 & V2.2 (19.01)            & Provides a complete bottom-up
  simulation platform, including ADAS and automation systems.                                                                     \\
SUMO \cite{SUMO}                               & V1.15.0 (22.11)           & Provides a purely microscopic traffic model that can be defined to customize each vehicle.                              \\
TORCS                                               & V1.3.8 (17.03)          & Support for running a large
  number of agents at the same time, allowing for scheduling functions in dense
  vehicle areas.                                    \\
SVL
  Simulator \cite{SVL}                                     & V2021.3 (21.05)         & Enables developers to simulate
  billions of miles and arbitrary corner cases to accelerate algorithm
  development and system integration.                     \\
V-Rep                                               & V3.6.2 (19.01)          & With a driving actions
  assessment function, which indicates the agent behavior based on the result.                                                          \\
CarMaker                                             & V10.0 (21.10)           & Specifically designed for the
  development and seamless testing of cars and light-duty vehicles in all
  development stages.                                   \\
CARLA \cite{carla}                                               & V0.9.13 (21.11)         & Various city maps are provided
  for autonomous driving algorithms, as well as support for customized sensor
  types and weather conditions.                    \\
AriSim \cite{airsim}                                              & V1.8.1 (22.06)          & The capability to quickly
  complete autonomous driving tests, and build various scenarios (urban,
  countryside, highway, field, etc.)                      \\

Apollo \cite{apolloscape}                                             & V8.0 (22.12)            & Support for learning and
  validation of single and multi-vehicle autonomous driving algorithms on urban
  scenarios.                                           \\
Autoware \cite{autoware}                                            & V1.11.0 (21.05)         & An open-source autonomous
  driving platform, which include all component of autonomous function for
  intelligent vehicle.                                     \\
Drive
  Constellation                               & V6.05 (22.10)           & Provides a computing platform
  based on two different servers that can undertake large-scale vehicle data
  interaction services.                              \\
MetaDrive \cite{metadrive}                                           & V0.2.6.0 (22.11)        & A wide range of road segments
  are available, which can be customized to generate a variety of complex
  scenarios, more suitable for reinforcement learning.  \\
\hline
\end{tabular}
\end{table*}

\subsection{Physical Platform}

With the increase in computer computing capabilities, simulation testing has become increasingly capable of meeting the testing requirements for various scenarios and has proven effective in solving the long-tail problem associated with such systems. However, pre-trained models used in a simulator typically require fine-tuning prior to implementation in the real world.  Moreover, while simulation testing can cover a wide range of scenarios, it can't account for all corner cases. Consequently, a professional and safe semi-open autonomous driving validation site is essential [\cite{LiComp}.

Autonomous driving technology achieved significant development over the past few decades, and several countries adopt policies permitting the testing of robotic taxis on public roads.  In the United States, Waymo is now permitted to test robotaxis on the streets of San Francisco from 2022. Nuro recently begins to deploy autonomous delivery vehicles in Arizona, California, and Texas. In England, Aurrigo is conducting a trial of an autonomous shuttle at Birmingham airport. Wayve is authorized to test autonomous vehicles over long distances between five cities. In China, the commercialization of autonomous driving is rapidly progressing, with companies such as Apollo, Pony, and Momenta already implementing IVs in several cities. Additionally, Waytous is working on unmanned transport in unstructured and closed scenarios and has already provided driverless solutions for several open-pit mines.

\section{Challenges and Future Perspectives \label{future}}
Considerable milestones have been achieved in autonomous driving, as evidenced by its successful validation on semi-open roads in various cities. However, its complete commercial deployment is yet to be realized due to numerous obstacles and impending challenges that need to be surmounted.
\subsection{Challenges}

The challenges in IVs are summarised below:
\begin{itemize}
     \item [1)] \textcolor{black}{Perception: autonomous driving frameworks heavily rely on perception data, however, most sensors are vulnerable to environmental effects and suffer from partial perception issues. As a result, potential hazards may be ignored, and these drawbacks present security challenges for autonomous driving.}
     \item [2)] \textcolor{black}{Planning: both pipeline and end-to-end planning have intrinsic limitations, and ensuring the production of high-quality outputs under uncertain and complex scenarios is an indispensable research objective.}
     \item [3)] \textcolor{black}{Safety: hacking for autonomous driving systems is increasing, even minor disruptions have possibly triggered significant deviations. Therefore, the deployment of autonomous driving methods on a massive scale necessitates robust measures to counter adversarial attacks.}
     \item[4)] \textcolor{black}{Dataset: simulators are essential for training and testing autonomous driving models, however, models well-trained in virtual environments often cannot be directly implemented in reality \cite{Proceeding}. Thus, bridging the gap between virtual and real data is imperative for advancing research in this field.}
\end{itemize}




\subsection{Future Perspectives}

The mechanism of the end-to-end planner is the closest to the human driver, according to the input state to calculate the output space. However, due to challenges in data, interpretability, generalization, and policies, end-to-end planners are still scarcely implemented in the real world. Herein, we propose some future perspectives in the field of end-to-end planning.

\begin{itemize}

  \item [$\bullet$] 
    \textcolor{black}{Interpretability: Machine learning receives criticism due to its black-box properties. The current intermediate feature representations are insufficient to explain the causality of its inference process. In the case of IV, the consequences of lacking interpretability could be catastrophic. Thus, providing clear and understandable interpretations for the motion planner is crucial in enhancing trust in intelligent vehicles (IVs). Moreover, this approach could assist in predicting and rectifying potential issues that may jeopardize the safety of the passengers.}
  \item [$\bullet$]
    \textcolor{black}{ Sim2Real: The simulation and the real environment have obvious differences in scenario diversity and environment complexity, making it challenging to align simulation data with real data \cite{SE1, SE2}. Consequently, the well-trained models in simulators may not optimally perform in real settings. Developing a model to bridge the gap between simulated and real environments is critical to address the challenges about data diversity and fairness, which is also a crucial research direction in end-to-end planning.}

  \item [$\bullet$]

    \textcolor{black}{Reliability: One critical bottleneck that impedes the development and deployment of IVs is the prohibitively high economic and time costs required to validate their reliability. Constructing an artificial-intelligence-based algorithm that can identify the corner cases in a short time is a key direction for the validation of IVs.}
  \item [$\bullet$]
    \textcolor{black}{Governance: IV is not only a technical issue, the sound policy is also crucial. Designing a framework that includes safety standards, data privacy regulations, and ethical guidelines is necessary to govern the development and deployment of IVs. This framework will promote accountability and transparency, reduce risks, and ensure that the public interest is defended.}

    

\end{itemize}

\bibliographystyle{IEEEtran}
\bibliography{Mylib.bib}

\begin{IEEEbiography}[{\includegraphics[width=1in,height=1.25in,clip,keepaspectratio]{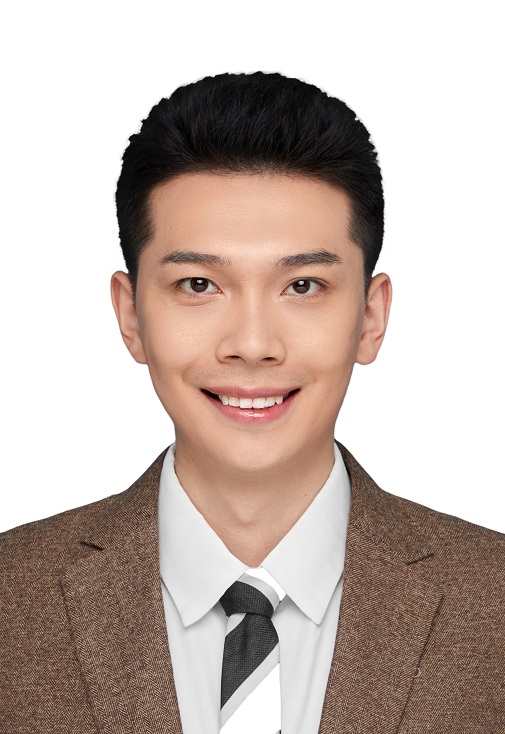}}] {Siyu Teng} received M.S. degree from Jilin University in 2021. Now he is a PhD Student at Department of Computer Science, Hong Kong Baptist University. His main interests are parallel planning, end-to-end autonomous driving and interpretable deep learning.
\end{IEEEbiography}

\begin{IEEEbiography}[{\includegraphics[width=1in,height=1.25in,clip,keepaspectratio]{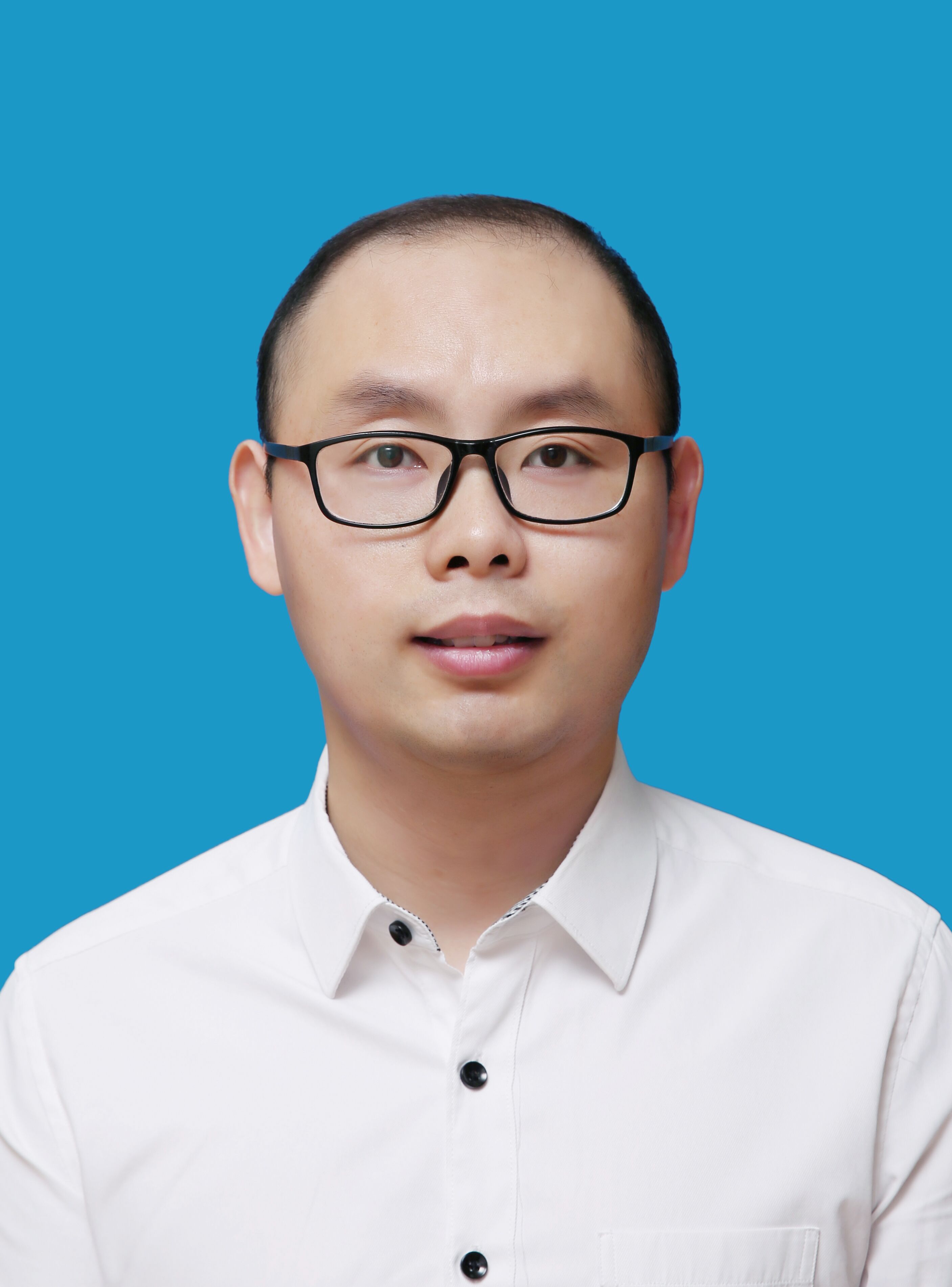}}]{Xuemin Hu} is currently an Associate Professor with School of Artificial Intelligence, Hubei University, Wuhan, China. He received the B.S. degree from Huazhong University of Science and Technology and the Ph.D. degree from Wuhan University in 2007 and in 2012, respectively. He was a visiting scholar in the University of Rhode Island, Kingston, RI, US from November 2015 to May 2016. His areas of interest include computer vision, machine learning, motion planning, and autonomous driving.
\end{IEEEbiography}

\begin{IEEEbiography}[{\includegraphics[width=1in,height=1.25in,clip,keepaspectratio]{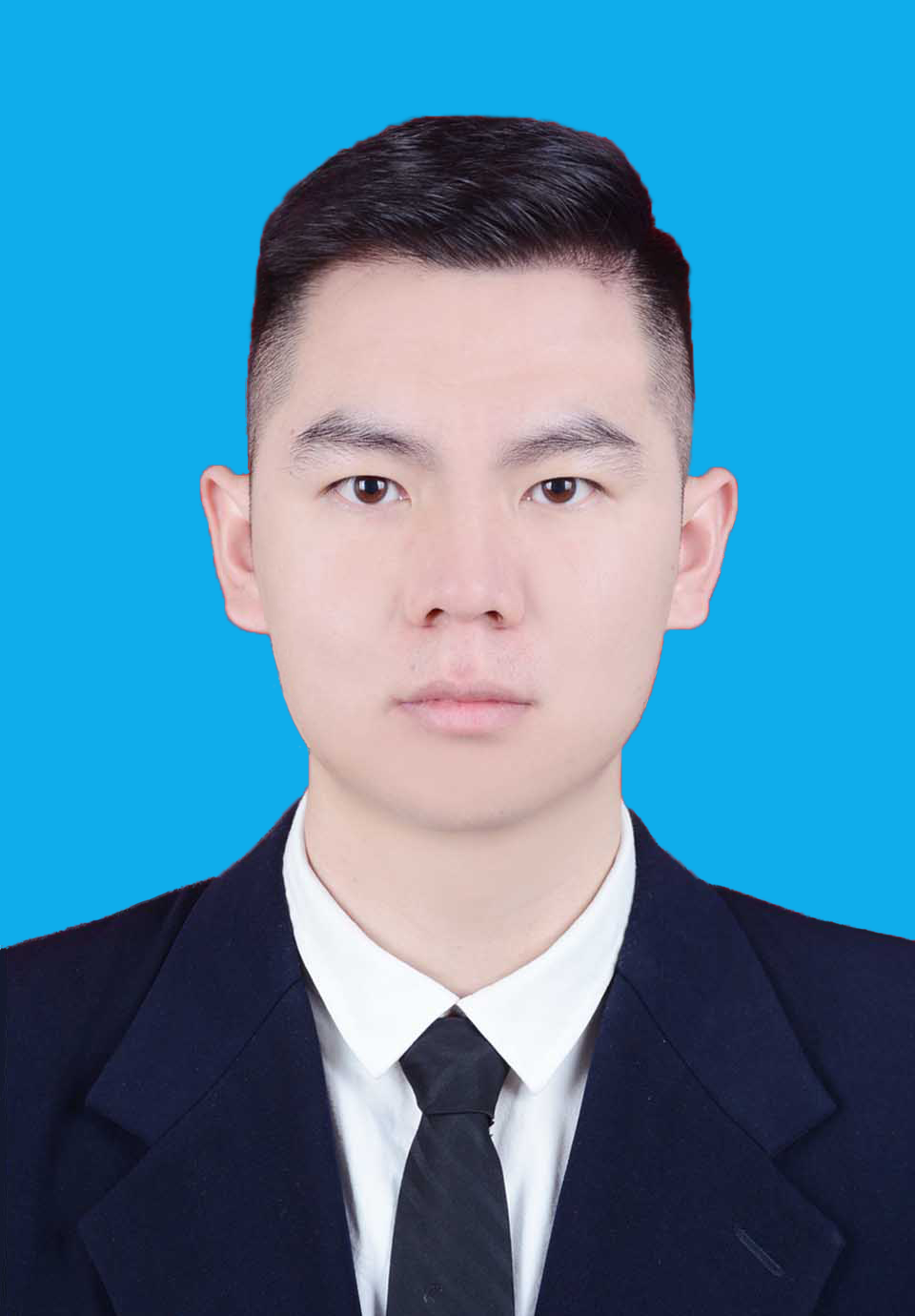}}]{Peng Deng} received the B.E. degree in vehicle engineering from China Agricultural University, Beijing, China. He is currently pursuing the M.S. degree with the School of Artificial Intelligence, Hubei University, Wuhan, China. His areas of interest include reinforcement learning and autonomous driving.
\end{IEEEbiography}

\begin{IEEEbiography}[{\includegraphics[width=1in,height=1.25in,clip,keepaspectratio]{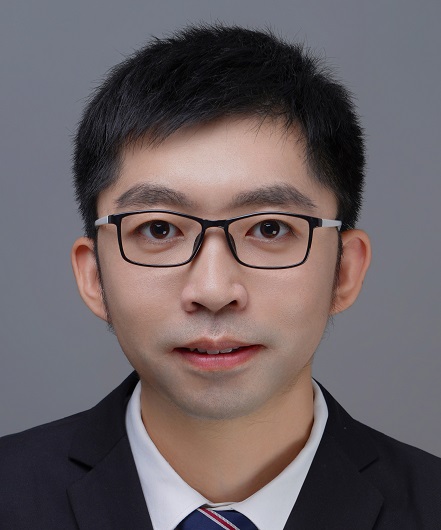}}]{Bai Li} (SM’13–M’18) received his B.S. degree in 2013 from the School of Advanced Engineering, Beihang University, China, and his Ph.D. degree in 2018 from the College of Control Science and Engineering, Zhejiang University, China. From Nov. 2016 to June 2017, he visited the Department of Civil and Environmental Engineering, University of Michigan (Ann Arbor), USA, as a joint training Ph.D. student. He is currently an associate professor in Hunan University. Before teaching at Hunan University, he worked in JDX R\&D Center of Automated Driving, JD Inc., China from 2018 to 2020 as an algorithm engineer. Prof. Li has been the first author of more than 70 journal/conference papers and two books related to numerical optimization, motion planning, and robotics. He was a recipient of the International Federation of Automatic Control (IFAC) 2014–2016 Best Journal Paper Prize from Engineering Applications of Artificial Intelligence. He is currently an Associate Editor of IEEE TRANSACTIONS ON INTELLIGENT VEHICLES. He was a recipient of the 2022 TIV Best Associate Editor Award. His research interest is rule-based motion planning methods for IVs.
\end{IEEEbiography}

\begin{IEEEbiography}[{\includegraphics[width=1in,height=1.25in,clip,keepaspectratio]{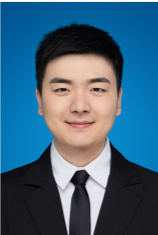}}]{Yuchen Li} received the B.E. degree from the University of Science and Technology Beijing in 2016, and the M.E. degrees from Beihang University in 2020. He is pursuing the Ph.D. degree in Hong Kong Baptist University. He is an intern at Waytous. His research interest covers computer vision, 3D object detection, and autonomous driving.
\end{IEEEbiography}

\begin{IEEEbiography}[{\includegraphics[width=1in,height=1.25in,clip,keepaspectratio]{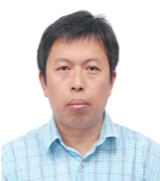}}]{Yunfeng Ai} received the Ph.D. degree from the University of Chinese Academy of Sciences, Beijing, China in 2006. He is Associate Professor at University of Chinese Academy of Sciences. He was a research fellow at Carnegie Mellon University. His current research interest covers computer vision, machine learning, robots, and autonomous driving.
\end{IEEEbiography}

\begin{IEEEbiography}[{\includegraphics[width=1in,height=1.25in,clip,keepaspectratio]{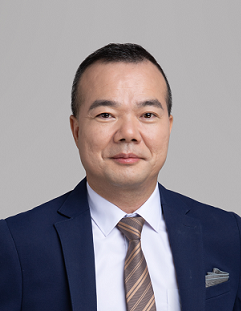}}]{Dongsheng Yang} received the Ph.D. degree in information system engineering from the National University of Defense Technology, Changsha, China, in 2004. He is currently a Professor with the School of Public Management/Emergency Management (The Laboratory for Military- Civilian Integration Emergency Command and Control), Jinan University, Guangzhou, China. His research interests include intelligent emergency response of complex systems, multiscale emergency command and control mode and mechanism, and parallel intelligent technology of emergency management.
\end{IEEEbiography}

\begin{IEEEbiography}[{\includegraphics[width=1in,height=1.25in,clip,keepaspectratio]{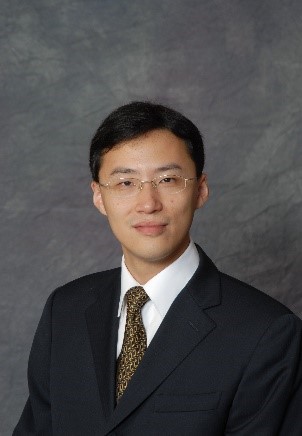}}]{Lingxi Li}  (S’04-M’08-SM’13) is currently a full professor in the Department of Electrical and Computer Engineering at Purdue School of Engineering and Technology, Indiana University-Purdue University Indianapolis (IUPUI), USA. Dr. Li received his Ph.D. degree in Electrical and Computer Engineering from the University of Illinois at Urbana-Champaign in 2008. Dr. Li’s current research focuses on modeling, analysis, control, and optimization of complex systems, connected and automated vehicles, intelligent transportation systems, digital twins and parallel intelligence, and human-machine interaction. He has authored/co-authored one book and over 130 research articles in refereed journals and conferences. Dr. Li was the recipient of five best paper awards, 2021 IEEE ITSS outstanding application award, 2017 outstanding research contributions award, 2012 T-ITS outstanding editorial service award, and several university research/teaching awards. He is currently serving as an associate editor for five international journals and has served as the General Chair, Program Chair, Program Co-Chair, etc., for 20+ international conferences. 
\end{IEEEbiography}

\begin{IEEEbiography}[{\includegraphics[width=1in,height=1.25in,clip,keepaspectratio]{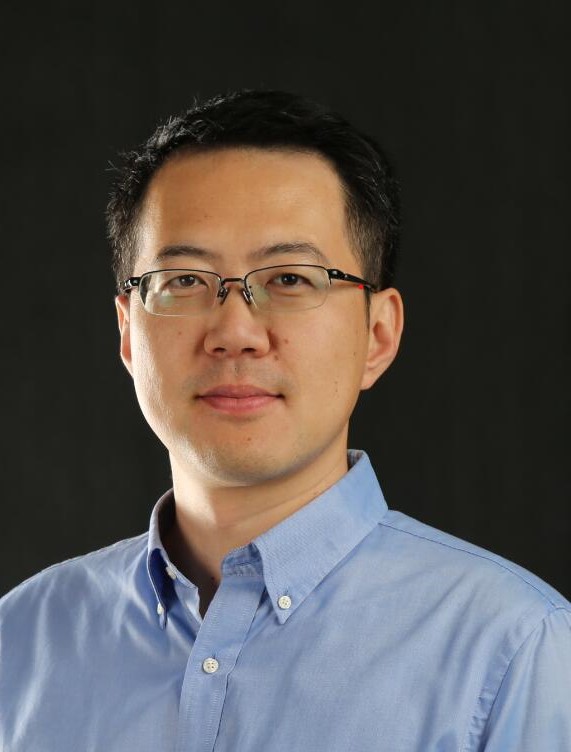}}]{Zhe XuanYuan} received the B.S. degree in electronic engineering from Peking University, Beijing, China, in 2005, and the Ph.D. degree in electronic and computer engineering from the Hong Kong University of Science and Technology, Hong Kong, in 2012. He is now an Associate Professor of Data Science with Beijing Normal University-Hong Kong Baptist University United International College, Zhuhai, China. His research interests include robot mapping and navigation, autonomous driving, and vehicular networks.
\end{IEEEbiography}

\begin{IEEEbiography}[{\includegraphics[width=1in,height=1.25in,clip,keepaspectratio]{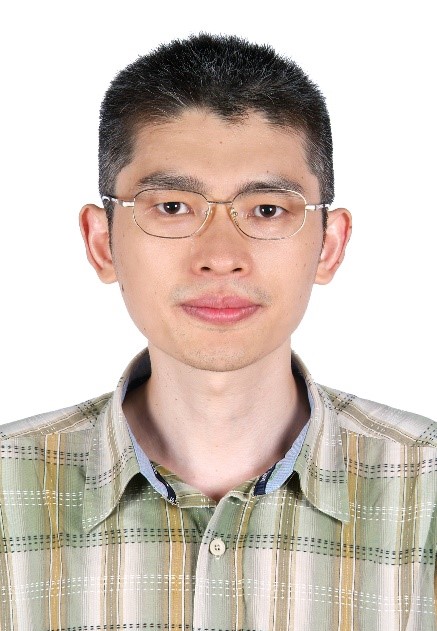}}]{Fenghua Zhu} (Senior Member, IEEE) received the Ph.D. degree in control theory and control engineering from the Institute of Automation, Chinese Academy of Sciences, Beijing, China, in 2008. He is currently an Associate Professor with the State Key Laboratory of Multimodal Artificial Intelligence Systems, China. His research interests include artificial transportation systems and parallel transportation management systems.
\end{IEEEbiography}

\begin{IEEEbiography}[{\includegraphics[width=1in,height=1.25in,clip,keepaspectratio]{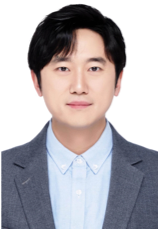}}]{Long Chen} (Senior Member, IEEE) received the Ph.D. degree from Wuhan University in 2013, he is currently a Professor with State Key Laboratory of Management and Control for Complex Systems, Institute of Automation, Chinese Academy of Sciences, Beijing, China. His research interests include autonomous driving, robotics, and artificial intelligence, where he has contributed more than 100 publications. He serves as an Associate Editor for the IEEE Transaction on Intelligent Transportation Systems, the IEEE/CAA Journal of Automatica Sinica, the IEEE Transaction on Intelligent Vehicle and the IEEE Technical Committee on Cyber-Physical Systems.
\end{IEEEbiography}

\end{document}